\documentclass[sigconf]{acmart}
\usepackage{multirow}
\usepackage{subcaption}
\usepackage{makecell}
\usepackage{soul}
\usepackage{colortbl}
\usepackage{enumitem}
\AtBeginDocument{%
  }

\newenvironment{tablenotes}[1][\scriptsize]{%
  #1
  \begin{list}{}{
    \topsep=1pt
    \partopsep=0pt
    \leftmargin=1.5em 
    \labelwidth=1em
    \labelsep=0.5em
  }%
}{%
  \end{list}%
}

\acmConference[ICSE '26]{International Conference
on Software Engineering}{April 12--18,
  2026}{Rio de Janeiro, Brazil}

\acmISBN{978-1-4503-XXXX-X/2018/06}

\newcounter{finding}
\newcommand{\finding}[1]{\refstepcounter{finding}
    \vspace{0.5mm}
    \begingroup
    \arrayrulewidth=3pt 
    \arrayrulecolor{gray} 
    \hspace*{-10pt}
    \renewcommand{\arraystretch}{1.2} 
    \begin{tabular}{|>{\cellcolor{gray!15}\hspace{0pt}\bfseries}p{\dimexpr\linewidth-3pt-2\tabcolsep}@{}}
        Finding \arabic{finding}: \normalfont #1 
    \end{tabular}
    \endgroup
    \vspace{0.5mm}
}

\begin{document}

\title{A Comprehensive Study of Deep Learning Model Fixing Approaches}


\author{Hanmo You}
\email{youhanmo@tju.edu.cn}
\orcid{0000-0003-2870-7801}
\affiliation{%
  \institution{College of Intelligence and Computing, Tianjin University}
  \city{Tianjin}
  \country{China}
}

\author{Zan Wang}
\email{wangzan@tju.edu.cn}
\authornotemark[1]
\orcid{0000-0001-6173-8170}
\affiliation{%
  \institution{College of Intelligence and Computing, Tianjin University}
  \city{Tianjin}
  \country{China}
}

\author{Zishuo Dong}
\email{zishuodong@tju.edu.cn}
\orcid{0009-0008-0782-5044}
\affiliation{%
  \institution{College of Intelligence and Computing, Tianjin University}
  \city{Tianjin}
  \country{China}
}

\author{Luanqi Mo}
\email{luanqimo@tju.edu.cn}
\orcid{0009-0008-7195-9899}
\affiliation{%
  \institution{College of Intelligence and Computing, Tianjin University}
  \city{Tianjin}
  \country{China}
}

\author{Jianjun Zhao}
\email{zhao@ait.kyushu-u.ac.jp}
\orcid{0000-0001-8083-4352}
\affiliation{%
  \institution{Kyushu University}
  \city{Fukuoka}
  \country{Japan}
}

\author{Junjie Chen}
\email{junjiechen@tju.edu.cn}
\authornote{Corresponding Authors}
\orcid{0000-0003-3056-9962}
\affiliation{%
  \institution{College of Intelligence and Computing, Tianjin University}
  \city{Tianjin}
  \country{China}
}








\renewcommand{\shortauthors}{You et al.}

\begin{abstract}
Deep Learning (DL) has been widely adopted in diverse industrial domains, including autonomous driving, intelligent healthcare, and aided programming. Like traditional software, DL systems are also prone to faults, whose malfunctioning may expose users to significant risks. Consequently, numerous approaches have been proposed to address these issues.
In this paper, we conduct a large-scale empirical study on 16 state-of-the-art DL model fixing approaches, spanning model-level, layer-level, and neuron-level categories, to comprehensively evaluate their performance. We assess not only their fixing effectiveness (their primary purpose) but also their impact on other critical properties, such as robustness, fairness, and backward compatibility. To ensure comprehensive and fair evaluation, we employ a diverse set of datasets, model architectures, and application domains within a uniform experimental setup for experimentation.
We summarize several key findings with implications for both industry and academia. For example, model-level approaches demonstrate superior fixing effectiveness compared to others. No single approach can achieve the best fixing performance while improving accuracy and maintaining all other properties. Thus, academia should prioritize research on mitigating these side effects. These insights highlight promising directions for future exploration in this field.

\end{abstract}





\copyrightyear{2026}
\acmYear{2026}
\setcopyright{rightsretained}
\acmConference[ICSE '26]{2026 IEEE/ACM 48th International Conference on Software Engineering}{April 12--18, 2026}{Rio de Janeiro, Brazil}
\acmBooktitle{2026 IEEE/ACM 48th International Conference on Software Engineering (ICSE '26), April 12--18, 2026, Rio de Janeiro, Brazil}
\acmPrice{}
\acmDOI{10.1145/3744916.3764528}
\acmISBN{979-8-4007-2025-3/26/04}

\maketitle

\section{Introduction}
\label{sec:introduction}

Deep learning (DL) has found extensive applications in a diverse range of domains, including autonomous driving~\cite{zhu2021survey}, intelligent healthcare~\cite{tozuka2023improvement}, and aided programming~\cite{DBLP:journals/tosem/HuangCJLYL24,DBLP:conf/emnlp/KangW00Y21,DBLP:journals/chinaf/ChenHHJJJJLLLLMPSSWWWWX25,rlforse}. Like traditional software, DL systems (mainly referring to DL models) are also prone to faults, which can lead to severe consequences. For example, in 2024 in Seattle, a Tesla vehicle operating in \textit{Full Self-Driving} mode caused several crashes and killed a motorcyclist~\cite{news1}. 
Therefore, fixing the faults in DL systems is essential to ensure the safety and reliability of relevant applications~\cite{AdRep,MODE,Apricot}.

To address these faults, researchers have proposed various DL model fixing approaches. 
In general, they can be classified into three categories based on the granularities they operate at, i.e., model-level, layer-level, and neuron-level approaches~\cite{fixing2023survey}.
Specifically, model-level approaches~\cite{Apricot,ENN,SpecRepair,HUDD,FSCMix,DeepRepair,HybridRepair} assume that faults originate from data limitations and perform retraining or finetuning on augmented~\cite{FSCMix} or selected datasets~\cite{HUDD} for fixing.
Layer-level approaches~\cite{MVDNN,3M-DNN,PRDNN,REASSURE,APRNN,PatchNAS} assume that poorly learned layers or sub-optimal structures may cause faulty behaviors, and target fixing these specific network layers or model architectures. 
Verification-guided~\cite{APRNN} or constraint-solving~\cite{REASSURE} methods are commonly used to enhance repair effectiveness. 
Neuron-level approaches posit that ill-trained neurons are responsible for faulty behaviors. They often localize these neurons and then apply search-based~\cite{Arachne} or optimization~\cite{I-REPAIR} methods to fix their weights.

Despite these advancements in academia, a recent interview study pointed out that these approaches are not adopted in industry due to two critical concerns~\cite{interview}.
First, practitioners do not have enough confidence in the effectiveness of these approaches because they were often evaluated on different datasets under different experimental setups, even under nonstandard setups.
For example, some approaches directly use part of the test-set data for fixing or adopt the test set (rather than the validation set) for evaluation~\cite{FSCMix,DeepRepair}, which is likely to cause data leakage or overfitting.
Second, practitioners are afraid that these approaches overly seek model correctness but overlook other essential properties of DL models, such as robustness~\cite{zhang22testing}, fairness~\cite{DBLP:conf/sigsoft/BrunM18}, and backward compatibility~\cite{DBLP:conf/kdd/SrivastavaNKSH20}.
Damaging these properties may cause significant consequences, including instability under extreme conditions~\cite{DBLP:conf/icse/TianPJR18}, discriminatory decisions~\cite{angwin2022machine}, or operational errors in human-DL interaction~\cite{you2025mitigating}. 
These concerns have been well-founded, but a comprehensive study is lacking to address them systematically.
Such a study is essential as it can offer an in-depth analysis and guidance on how to use these approaches more effectively in practice.

To bridge the gap between academia and industry, we conducted the first study to investigate the performance of these DL model fixing approaches comprehensively. 
(1) We studied \textbf{16} state-of-the-art or typical DL model fixing approaches, covering all three categories (i.e., model-level, layer-level, and neuron-level). 
(2) We focused on both the \textbf{correctness} (the primary purpose of DL model fixing) and the other three critical properties for DL models, i.e., \textbf{robustness}, \textbf{fairness}, and \textbf{backward compatibility}, during the evaluation of these approaches.
As suggested by the latest survey~\cite{interview}, we also considered the \textbf{applicability} and \textbf{efficiency} of these approaches, which are also of concern to some practitioners.
(3) We adopted \textbf{a diverse set of subjects} for sufficient evaluation, covering different datasets, model structures, and application domains.
Particularly, we designed a \textbf{uniform and reasonable experimental setup} for all the studied approaches.

From our comprehensive study, we obtained a series of findings and implications that can help guide future research and practical use of DL model fixing approaches. 
We briefly introduce some key findings and implications as follows: 

\begin{enumerate}[leftmargin=*]
\item Under the correctness metric, model-level approaches outperform layer-level ones in accuracy improvement, which in turn surpass neuron-level ones; for error correction, model-level approaches are superior to neuron-level ones, which outperform layer-level ones. Overall, model-level approaches have superior effectiveness in correctness improvement.

\item  Current fixing approaches are more effective for small-scale models. As the model complexity increases, their performance degrades. Therefore, there is an urgent need for more effective fixing approaches for large and complex DL models, such as large language models (LLMs).

\item  No single approach can achieve superior fixing performance while improving accuracy and maintaining fairness, robustness, and backward compatibility. Hence, future research should focus on mitigating these side effects. Post-processing methods present a promising solution.

\end{enumerate}

To sum up, our work makes the following major contributions:

\begin{itemize}[leftmargin=*]
    \item We conducted \textbf{the first large-scale empirical study} to comprehensively evaluate the performance of 16 state-of-the-art approaches that aim to improve the correctness of DL models.
    \item We summarized \textbf{a set of findings and implications} by systematically analyzing and comparing the performance of different approaches under a uniform experimental setup. 
    \item We re-implemented some of the approaches and established \textbf{a uniform evaluation platform} for DL fixing approaches. This platform can facilitate comparison for future research in this area. Moreover, we have made all our experimental data publicly available on our homepage~\cite{homepage}.
\end{itemize}

\section{BACKGROUND AND RELATED WORK}

\subsection{Important Properties of DL Models}
\label{sec:formalization}
In this section, we define the concepts used in this paper. Let $\mathcal{M}$ be the DL model under test, and $x$ and $y$ be a data point and its ground truth label belonging to a future unknown data distribution $\mathcal{D}$. $p_\mathcal{M}(x)$ is the predicted label of $x$.

\begin{description}[leftmargin=0pt]

\item[Correctness] is the fundamental property of DL models, which measures whether DL models make correct predictions.
The model correctness $E(\mathcal{M})$ can be formalized as the probability that $p_\mathcal{M}(x)$ and $y$ are identical.
\begin{equation}
E(\mathcal{M}) = \mathbb{E}_{x \sim \mathcal{D}} [ p_\mathcal{M}(x) = y ]    
\end{equation}

\end{description}

In addition to correctness, developers put much effort into maintaining some other important properties~\cite{zhang22testing,you2025mitigating,DBLP:conf/issta/Pham024} of DL models. The most commonly mentioned are robustness, fairness, and backward compatibility~\cite{zhang22testing,you2025mitigating}.

\begin{description}[leftmargin=0pt]
\item[Robustness] is defined as the resilience of a DL system's correctness in the face of small perturbations ~\cite{zhang22testing,INNER}. For example, adversarial examples~\cite{DBLP:conf/icse/ZhouLKGZ0Z020} and environmental noise (e.g., fog, rain)~\cite{DBLP:conf/icse/TianPJR18} may influence the decision-making of DL-based autonomous driving systems. This susceptibility underscores the crucial importance of robustness in DL models, which is vital for their reliable operation. Suppose that $\Delta$ is the perturbation bounded by the norm $||\Delta||\leq \epsilon$, robustness $R(\mathcal{M})$ can be formalized as:
\begin{equation}
R(\mathcal{M}) = \mathbb{E}_{x \sim \mathcal{D}} [ p_\mathcal{M}(x + \Delta) = y | \ ||\Delta||\leq \epsilon ]
\end{equation}
\end{description}

\begin{description}[leftmargin=0pt]
\item[Fairness] is a non-functional software property that should be treated as a first-class entity throughout the SE process~\cite{DBLP:conf/sigsoft/BrunM18}. Fairness definitions~\cite{DBLP:journals/tosem/ChenZHHS24} mainly fall into individual and group fairness. The former requires that software generate similar predictive results for similar individuals, while the latter demands that software treat different demographic groups in a similar way.
Fairness problems are also common in DL models; their statistical nature means they can easily learn biases from the collected data, leading to unfairness problems~\cite{zhang22testing}. For example, the software used by the US courts to assess the risk of a criminal re-offending when making bail and sentencing decisions has been shown to exhibit racial bias~\cite{angwin2022machine}.
Therefore, ensuring the fairness of DL models is important.
Suppose that $x_a$ and $y_a$ are a data point and its corresponding ground truth label from $\mathcal{D}_a$ (which is the set of data with protected attribute $a$); then the group fairness $F(\mathcal{M})$ can be formalized as:
\begin{equation}
F(\mathcal{M}) = |\mathbb{E}_{x \sim \mathcal{D}_a}[ p_\mathcal{M}(x_a) = y_a] - \mathbb{E}_{x \sim \mathcal{D}}[ p_\mathcal{M}(x) = y]|
\end{equation}
\end{description}

\begin{description}[leftmargin=0pt]
\item[Backward Compatibility] means that after model evolution, the updated models that are more
accurate may still break human expectations since they may introduce regression faults that are not present in earlier versions of the models~\cite{DBLP:conf/kdd/SrivastavaNKSH20,DBLP:conf/icse/YouWCLL23}. A DL model lacking backward compatibility in safety-critical domains may cause significant risks. For example, doctors may be accustomed to the previous model in a DL-based disease diagnosis system. When regression faults occur in the updated model, they may be unfamiliar with these faults and rely on their past experience, leading to misdiagnosis~\cite{you2025mitigating}. 
Previous work~\cite{DBLP:conf/icse/YouWCLL23,DBLP:conf/sigsoft/You25} has shown that the DL model evolution process, including fixing, can lead to many regression faults. Hence, we regard backward compatibility as a crucial property worthy of investigation in our study. Suppose $\mathcal{M}^\prime$ is the fixed model, the Backward Compatibility $B(\mathcal{M})$ can be formalized as:
\begin{equation}
B(\mathcal{M}) = \mathbb{E}_{x \sim \mathcal{D}}[ p_\mathcal{M^\prime}(x) \neq y|p_\mathcal{M}(x) = y]
\end{equation}
\end{description}

\subsection{Model Fixing for Correctness Improvement}

Recently, researchers have proposed a number of approaches to fixing DL models. Following the existing work~\cite{fixing2023survey,BIRDNN}, we define \textit{DL Model Fixing} as follows:

Given a DL model to be fixed $\mathcal{M}$, and a set of mispredictions $\{\Phi_i\}_{i=1}^\phi$, the fixing process is to rectify $\mathcal{M}$ to $\mathcal{M}^{\prime}$ by adjusting improper data for re-training or directly correcting ill-trained layers or neurons in $\mathcal{M}$, trying to make $\mathcal{M}^{\prime}$ perform correctly on $\{\Phi_i\}_{i=1}^\phi$.
Note that the Artificial Intelligence (AI) community also develops new training algorithms or architectures for improving DL models. These contributions are different from those made by model fixing in the Software Engineering (SE) community, where the latter targets diagnosing potential root causes and fixing faulty elements (such as data or neurons) to improve model correctness.
In theory, new training algorithms and architectures can be combined with model fixing approaches for further enhancement.
Hence, this study focused on these model fixing approaches from the SE perspective and will explore their combination with those AI contributions in the future.

To perform a comprehensive study of DL model fixing approaches, we first conducted a literature review on this topic from prominent conferences and journals, including ICSE, FSE, ASE, ISSTA, TOSEM, ICLR, AAAI, etc. 
We used keywords like \textbf{repair}, \textbf{debug}, and \textbf{fix} to identify the papers likely relevant to improving DL models' correctness from the SE perspective.

Each paper was then manually checked to ensure its relevance.
Following recommendations from existing work~\cite{fixing2023survey}, we excluded studies that could lead to conceptual ambiguity. For example, tools like AutoTrainer~\cite{DBLP:conf/icse/ZhangZMS21}, which focus on debugging training code, are categorized as \textit{DL program repair }(i.e., modifying source code) rather than \textit{DL model repair} (i.e., adjusting model weights), and thus fall outside the scope of our study. Please note that the primary focus of the approaches we selected is on correctness-related issues. That is, they consistently define faults as incorrect model predictions. We acknowledge that many other studies target different types of issues, such as fairness-related bugs~\cite{DBLP:conf/issta/YangJS024} or vulnerabilities introduced by backdoors~\cite{DBLP:journals/ojcomps/LiZWS23}. Evaluating such diverse targets would require distinct metrics and benchmarks, and could dilute the focus of our study. Therefore, we concentrate on correctness-related faults, given their widespread relevance in real-world applications.
Table~\ref{tab:works} shows all the DL model fixing approaches we finally collected. 
\textbf{Among the 36 approaches, 72\% were published in SE or formalization conferences/journals}.

\section{STUDIED APPROACHES}
\label{sec:studied_approaches}
From Table~\ref{tab:works}, 
we further selected our studied approaches according to the following criteria.

\begin{description}[leftmargin=*]
    \item[Criterion 1] \textbf{(Latest Work):} To ensure the effectiveness and representativeness of our study, we prefer to select the methods proposed in recent five years (from 2019 to 2024) since they usually can achieve much better performance than the previous ones.
    
    \item[Criterion 2] \textbf{(Open-source or Reproducible):} To ensure the correctness and reliability of our study, we reproduced the results of the selected approaches before conducting our experiments. We selected the approaches that are open-source or can be replicated through our re-implementation with reported configuration, ignoring those that cannot be replicated or adapted. We are grateful to the authors who assisted us in re-implementing and reproducing their results.

\end{description}

Criterion 1 has filtered out one approach, and Criterion 2 has filtered out 19 approaches. Finally, we have selected 16 state-of-the-art fixing approaches, which cover all categories and are bolded in Table~\ref{tab:works}. The following section overviews these approaches, with key strategies underlined for clarity.

\begin{table}[t]

\centering
\caption{Summary of Fixing Approaches}
\label{tab:works}
\begin{tabular}{c|l}
\toprule
\textbf{Category} & \textbf{Methods} \\ \midrule
Model-level       & \makecell[l]{\textbf{Apricot}~\cite{Apricot}, \textbf{Editable Training}~\cite{ENN}, \\
Veritex~\cite{Veritex}, SpecRepair~\cite{SpecRepair}, BIRDNN-RT~\cite{BIRDNN}, \\
\textbf{HUDD}~\cite{HUDD},
\textbf{FSGMix}~\cite{FSCMix}, \textbf{DeepRepair}~\cite{DeepRepair}, \\
\textbf{HybridRepair}~\cite{HybridRepair}, MetaRepair~\cite{MetaRepair}, SAFE~\cite{SAFE}} \\ 
\midrule
Layer-level       & \makecell[l]{MVDNN~\cite{MVDNN}, \textbf{3M-DNN}~\cite{3M-DNN}, ArchRepair~\cite{ArchRepair}, \\
PRDNN~\cite{PRDNN}, LRNN~\cite{LRNN},  \textbf{REASSURE}~\cite{REASSURE}, \\
RIPPLE~\cite{RIPPLE}, ARNN~\cite{ARNN}, APRNN~\cite{APRNN}, SR~\cite{SR}, \\
 PatchNAS~\cite{PatchNAS}}          \\
\midrule
Neuron-level      & \makecell[l]{MODE~\cite{MODE}, NREPAIR~\cite{NREPAIR}, \textbf{Arachne}~\cite{Arachne}, \\ \textbf{CARE}~\cite{CARE}, \textbf{NeuRecover}~\cite{NeuRecover}, BIRDNN-FT~\cite{BIRDNN}, \\ 
\textbf{I-REPAIR}~\cite{I-REPAIR}, \textbf{DISTRREP}~\cite{DISTRREP}, ADREP~\cite{AdRep}, \\
\textbf{GenMuNN}~\cite{GenMuNN}, NNREPAIR~\cite{NNREPAIR}, IDNN~\cite{IDNN}, \\
\textbf{VERE}~\cite{VERE}, \textbf{INNER}~\cite{INNER}}         \\
\bottomrule
\end{tabular}


\end{table}

\subsection{Model-level Fixing}

Model-level fixing is typically applied when developers identify flaws in training data (e.g., missing data types) or training processes (e.g., failing to learn certain types of data), thus making global adjustments more suitable. For data-related flaws, developers augment or select data to rectify the data distribution before finetuning. For process-related flaws, they focus on retraining to learn under-learned patterns. Based on our selection criteria, we analyze six representative model-level approaches.

\begin{itemize}[leftmargin=*]
\item  \textbf{Apricot(AP)}~\cite{Apricot} is a \ul{white-box} \ul{weight-adjustment} fixing approach. Its key insight is that training a model on different subsets of the training set can reveal how to adjust the model's weights to address misclassified test cases. Specifically, Apricot divides the training set into several subsets, using each subset to train a submodel. Then, it adjusts the weights of the model to be fixed, making them approach those of submodels with correct behaviors and differ from those of incorrect behaviors.

\item  \textbf{Editable Training(ET)}~\cite{ENN} is a \ul{retraining}-based approach whose insight is that training a DL model in a manner that the trained weights are easily editable afterward. ET designs a \ul{new objective function} that considers learning wrongly predicted samples while reducing the performance degradation brought by editing.

\item \textbf{FSGMix(FM)}~\cite{FSCMix} is a black-box approach using \ul{failure examples} to guide \ul{training data augmentation}. It collects failing samples via few-shot sampling to estimate the distribution of faulty samples to determine augmentation parameters. For each original training image, two augmented ones are generated, and the model is finetuned on this expanded dataset to reduce faulty behaviors.

\item \textbf{DeepRepair(DR)}~\cite{DeepRepair} is a {black-box}, \ul{style-guided data augmentation} fixing approach. It analyzes failure patterns in failing samples to construct a style-transfer model, generating similar failure-like samples for finetuning to fix such cases.

\item  \textbf{HUDD}~\cite{HUDD} is a \ul{white-box} fixing approach based on \ul{counterexample selection}. It posits that retraining with representative failing samples can facilitate fixing. HUDD clusters the internal heatmaps of failing samples to identify representative cases in each cluster, identified as root causes of failure. It selects cases similar to these representatives from the repair set for retraining.

\item  \textbf{HybridRepair(HR)}~\cite{HybridRepair} is an \ul{annotation-efficient data-selection} fixing approach to address insufficient data distribution. It annotates unlabeled new data to enhance training data density and uses these data to fix sparse regions during finetuning.

\end{itemize}

\subsection{Layer-level Fixing}


Layer-level fixing is applied when developers target specific poorly learned layers or structures causing faulty behaviors. These approaches aim to adjust weights or optimize structures using verification-guided or constraint-based methods. In contrast to model-level approaches, which globally modify all weights non-differentially, layer-level fixing focuses on preferential adjustments. However, some layer-level approaches depend on model-specific structures, limiting their applicability~\cite{ishimoto2024repairs}. We, therefore, selected two more adaptable layer-level approaches for studying.

\begin{itemize}[leftmargin=*]

\item  \textbf{3M-DNN(3M)}~\cite{3M-DNN} is a \ul{search}-based approach. Its core idea is splitting the model into sub-networks along separation layers. It tries to search for minimal changes for each sub-network to achieve the overall model fixing. Due to its high computational cost from using a safety-analysis \ul{verification} tool, we use its configuration to fix only one sub-network for each DL model.

\item  \textbf{REASSURE(RA)}~\cite{REASSURE}  is an \ul{optimization}-based approach designed to fix the continuous piecewise linear properties of ReLU networks. It synthesizes a patch network for the linear regions of failing samples, generates a support network, and \ul{solves a linear programming problem} to determine the optimal parameters for the patch network. Finally, it integrates these patches and support networks with the original model for fixing.

\end{itemize}

\subsection{Neuron-level Fixing}

Neuron-level approaches adjust the weights of specific misbehaving neurons responsible for faulty behaviors in DL models. These approaches are considered more efficient and less disruptive to other model properties compared to model- or layer-level fixes. They typically involve two steps: localizing faulty neurons and searching for optimal weight adjustment. We studied eight representative neuron-level approaches.

\begin{itemize}[leftmargin=*]

\item \textbf{Arachne(AR)}~\cite{Arachne} is a \ul{search}-based fixing approach. It localizes neural weights that affect negative samples more than positive ones for fixing. It first assesses the impact of each neural weight using \ul{gradient}-based loss for localization. Then, it uses differential evolution to search for the optimal weights for these neurons.

\item \textbf{CARE}~\cite{CARE} is a \ul{search}-based approach. Its core insight is capturing a neuron's causal influence on model performance for a given property. It first localizes faulty neurons via \ul{causality analysis} and then fixes their weights via searching. CARE was originally designed for backdoors/fairness bugs, and existing work~\cite{ishimoto2024repairs} has proven it applicable for fixing wrong predictions.

\item  \textbf{NeuRecover(NR)}~\cite{NeuRecover} is a \ul{search}-based approach. Its insight is to find a point in the \ul{training history} where the model correctly predicted a sample that is now mispredicted. Then, by comparing the past and current models, it identifies the neuron that can safely fix the faulty behaviors and search for optimal weights.

\item  \textbf{I-REPAIR(IR)}~\cite{I-REPAIR} is a \ul{focal-tuning}-based approach designed for fixing scenarios without access to the training set and with only limited access to certain samples. First, it estimates the influence of each neuron weight using \ul{gradients} to locate weights that have a more significant impact on failure samples than on correct ones. Then, it updates these weights using back-propagation.

\item  \textbf{DISTRREP(DTR)}~\cite{DISTRREP} is a \ul{search}-based approach. It considers the fact that there are more critical misclassifications that should be given higher priority to be fixed. First, it conducts distributed fault localization for each misclassification and then searches for patches. Next, it integrates these patches according to the \ul{risk levels} of these misclassifications to fix the incorrect behaviors.

\item  \textbf{GenMuNN(GM)}~\cite{GenMuNN} employs a \ul{mutation}-based genetic algorithm for fixing. First, it conducts \ul{spectrum analysis} on neuron weights. Then, based on the suspiciousness of these weights, it introduces mutations to generate model variants for fixing.

\item  \textbf{VERE}~\cite{VERE} is a \ul{verification}-based approach. It is guided by reachability analysis in the form of linear approximation from formal verification.
It first performs fault localization by quantifying the benefits of repairing a neuron based on \ul{model perturbation}. Then, it constructs an \ul{optimization} problem for fixing. 

\item  \textbf{INNER(IN)}~\cite{INNER} is an \ul{interpretability}-based approach via \ul{feature analysis}. It first uses a model probing method to evaluate each neuron's contribution to undesired behaviors. By analyzing influential features in hidden layers, INNER identifies faulty neurons causing unexpected behaviors. Then, it fixes them by aligning normal and abnormal features. It sets an \ul{objective function} to make neurons learn target features from failure samples to optimize their weights. 

\end{itemize}

\section{Research Questions}

In this study, we focus on the following research questions (RQs).
\begin{itemize}[leftmargin=*]

    \item \textbf{RQ1 (Applicability of fixing approaches):} \textit{How applicable are DL model fixing approaches across different datasets?} 

    \item \textbf{RQ2 (Effectiveness of fixing approaches):} \textit{What is the overall effectiveness of DL model fixing approaches on correctness?}

    \item \textbf{RQ3 (Impacts on other properties):} \textit{What impacts do DL model fixing approaches have on other important properties?}

    \item \textbf{RQ4 (Efficiency of fixing approaches):} \textit{What are the time costs of DL model fixing approaches?}
    
\end{itemize}

\smallskip
\noindent We design the RQs according to the limitations of existing works:

\begin{description}[leftmargin=*]
\item[Limitation 1] \textbf{(Incomplete Datasets/Models)}.
Most approaches have been evaluated on different datasets. For example, Apricot~\cite{Apricot} is evaluated only on the CIFAR10 dataset, GenMuNN~\cite{GenMuNN} only on the MNIST dataset, and DISTRREP~\cite{DISTRREP} only on a rarely used dataset, BDD100K. 
When it comes to the model structures for evaluation, they also vary greatly. Apricot~\cite{Apricot} focuses on CNN and ResNet models, while CARE~\cite{CARE} focuses on shallow-layer FFNN and CNN models. 
Notably, few approaches have been evaluated on a sufficiently diverse set of datasets, tasks, and model structures. This situation is insufficient for a comprehensive understanding of their applicability \textbf{(RQ1)}, effectiveness \textbf{(RQ2)}, and efficiency \textbf{(RQ4)}.

\item[Limitation 2] \textbf{(Inconsistent Evaluation Metrics)}.
Most fixing approaches aim to improve the correctness of DL models, and they concentrate on one or two metrics related to measuring fixing capabilities (e.g., accuracy) for evaluation~\cite{AIREPAIR}. Besides metrics related to correctness, some metrics related to other properties (e.g., robustness, fairness) are valuable in industrial practice but are seldom considered. Maintaining these properties during the fixing process is equally vital for ensuring the practical utility and safety of fixed models in deployment~\cite{DBLP:conf/issta/Pham024}.
The absence of evaluations on these properties hinders a comprehensive comparison of different approaches \textbf{(RQ3)}. 

\item[Limitation 3] \textbf{(Nonstandard Evaluation Processes)}.
We observe that different fixing approaches are evaluated on different experimental setups, which is common among those approaches relying on additional data to support fixing. Some works~\cite{DeepRepair,FSCMix} use the validation set/part of the test set data for fixing or directly use the test set for evaluation (rather than the validation set). These nonstandard evaluation processes can lead to data leakage, causing biased or unreliable evaluations on fixing effectiveness \textbf{(RQ2)}. 
\end{description}

\begin{description}[leftmargin=*]
\item[To Address Limitation 1.] We replicated the experimental results of existing approaches on their original datasets. Then, we extended the experiments to datasets and model structures selected by us, further evaluating their effectiveness.
\item[To Address Limitation 2.] We employed widely used performance metrics (i.e., two correctness metrics and three metrics related to other properties) to evaluate their effectiveness systematically.
\item[To Address Limitation 3.] We regularized the evaluation process to prevent potential data leakage; for those approaches that require additional data for fixing, we prepared a specific dataset for repair purposes instead of using validation/test sets~\cite{I-REPAIR}. We used the validation set to select models with optimal fixing performance and the test set for final evaluation, avoiding overfitting~\cite{DBLP:conf/kdd/LvDLCFHZJDT21}.
\end{description}
In this way, we can better understand the performance of existing approaches and gain valuable insights for more effective exploration of DL model fixing approaches.

\section{EXPERIMENTAL SETUP}
\label{sec:exmperiment}

\subsection{Benchmark}
\label{sec:benchmark}
In this study, we aim to conduct a comprehensive and uniform comparison among existing approaches. Specifically, we selected benchmarks according to the following criteria:

\begin{description}[leftmargin=*]
\item[Criterion 1] \textbf{(Widely-used):} To ensure study representativeness and ease the adaptation of the studied approaches to our experiment, we selected representative datasets and models that are commonly used for evaluating existing fixing approaches. Notably, all the studied works employed image classification as the benchmark modality and task, with their fixing approaches specifically designed or implemented for this setup; thus, we adopted the same configuration. Exploring the effectiveness of fixing approaches across different modalities and tasks is reserved for future work.

\item[Criterion 2] \textbf{(Diverse):} To draw comprehensive conclusions, we selected datasets and models with diverse characteristics. For datasets, we considered different application domains (i.e., digital, face, and object classification), classification category counts (i.e., 2 to 1000), and sensitive attributes (i.e., color and race). For models, we selected those with different structures (e.g., LeNet, VGG, ResNet, and DenseNet), scales (e.g., 8 to 249 layers), and training paradigms (i.e., pre-trained DenseNet121 and others trained from scratch).
\end{description}

Table~\ref{table:datasets} summarizes selected datasets, including model structures, performances, and the number of samples. MNIST~\cite{MNIST_dataset} is a digital-recognition dataset. CIFAR10~\cite{CIFAR10_dataset}, CIFAR10S~\cite{CIFAR10S_dataset}, and ImageNet~\cite{IMAGENET_dataset} are object-recognition datasets. CIFAR10S~\cite{CIFAR10S_dataset} and UTKFace~\cite{UTKFACE_dataset} are widely used in fairness-related tasks, so we use them to evaluate fairness following the existing study~\cite{DBLP:conf/issta/YangJS024}. CIFAR10S~\cite{CIFAR10S_dataset} is a variant of CIFAR10, containing both colored and non-colored images, with color being its sensitive attribute. UTKFace~\cite{UTKFACE_dataset} is a face-recognition dataset, and its sensitive attribute is race.
We select models that have satisfied performance as recommended in existing works~\cite{MNIST_dataset, CIFAR10_VGG16_use, 
DBLP:conf/issta/YangJS024,
UTKFACE_FaceNet_use, ArchRepair}. 
In addition, we present the number of samples for model training, validation, repair, and testing per dataset following existing works~\cite{I-REPAIR,DBLP:conf/nips/GhorbaniZ20}.

\begin{table}[t]
\centering
\caption{Summary of Adopted Benchmark Datasets}
\label{table:datasets}
\scalebox{0.65}{


\begin{tabular}{c|cccc|c}
\toprule
\textbf{Datasets} & \textbf{Model} & \textbf{\#Layers} & \textbf{\#Classes} & \textbf{Accuracy}  & \textbf{\#Train/\#Valid/\#Repair/\#Test}\\
\midrule
MNIST & LeNet5 & 8 & 10  & 97.90\%   & 54,000/6,000/5000/5000          \\
UTKface & FaceNet & 11 & 2 & 90.33\% & 17,778/1,975/1,976/1,976          \\
CIFAR10 & VGG16 & 42 & 10 & 86.22\%   & 45,000/5,000/5,000/5,000          \\
CIFAR10S & ResNet18 & 69 & 10 & 65.54\% &  45,000/5,000/10,000/10,000         \\
ImageNet  & DenseNet121 & 249 & 1,000 & 71.70\% & 1.2 M+/50,000/25,000/25,000         \\
\bottomrule
\end{tabular} 
}
\begin{tablenotes}
\scriptsize
\item[*] In this table, we sorted these datasets by model complexity. LeNet5 (1st row) has the fewest layers and simplest algorithm, while DenseNet121 (5th row) has the most layers and is the most complex.
\item[*] 1.2M+ indicates more than 1.2 million training images of ImageNet.
\end{tablenotes}


\end{table}

\subsection{Metrics}
\label{sec:metrics}

For presentation, we use the symbols consistent with Section~\ref{sec:formalization}. Additionally, let $N$ denote the number of test inputs, $x_i$ the $i$-th test input, and $y_i$ its ground truth from the test set. $\mathcal{I}(\cdot)$ is the indicator function, which returns 1 when the condition is true and 0 otherwise.

\begin{description}[leftmargin=*]
\item[Correctness]
To evaluate model fixing approaches, we employ two widely-used metrics from existing works~\cite{Arachne,NeuRecover}: Accuracy and Repair Rate (RR), defined by Formulas~\ref{eq:acc} and~\ref{eq:rr}, respectively. Specifically, the former metric focuses on the overall correctness of DL models, while the latter metric quantifies the fixing effectiveness on incorrectly predicted test inputs. For both Accuracy and RR, the larger the value, the better the performance of the fixing process.
\begin{equation}
	\label{eq:acc}
	\mathrm{\textit{Accuracy}} = \frac{1}{N} \sum_{i=1}^{N} \mathcal{I}(p_{\mathcal{M}^\prime}(x_i) = y_i)
\end{equation}
\begin{equation}
	\label{eq:rr}
	\mathrm{\textit{RR}} = \frac{1}{N} \sum_{i=1}^{N} \mathcal{I}(p_{\mathcal{M}}(x_i) \neq y_i,p_{\mathcal{M}^\prime}(x_i) = y_i)
\end{equation}
\item[Adversarial Robustness]
The robustness metric measures how well a model can withstand inputs with small perturbations. Various metrics for measuring robustness and Attack Success Rate (ASR) is commonly used to measure the model's robustness against adversarial examples~\cite{INNER}. As shown in Formula~\ref{eq:asr}, the ASR is defined as the number of adversarial examples successfully generated within the entire test set. Here, $x_i^{\prime}$ represents the adversarial example derived from $x_i$. The smaller the value, the more robust the DL model is.
\begin{equation}
	\label{eq:asr}
	\mathrm{\textit{ASR}} = \frac{1}{N} \sum_{i=1}^{N} \mathcal{I}(p_{\mathcal{M}^\prime}(x_i) = y_i,p_{\mathcal{M}^\prime}(x_i^{\prime}) \neq y_i)
\end{equation}

\item[Fairness] The fairness metric is designed to determine whether unprivileged and privileged groups are treated differently. Recent research~\cite{DBLP:conf/issta/YangJS024} has shown that the Average Absolute Odds Difference (AAOD) has the highest overall correlation with other fairness metrics and is thus more representative. Therefore, we select AAOD as the metric to measure fairness.
As shown in Formula~\ref{eq:AAOD}, AAOD is the average of the absolute difference in true/false positive rates between unprivileged and privileged groups. 
Here, $A$ is a sensitive (or protected) attribute: $A=1$  for the privileged and  $A=0$ for the unprivileged group. $Y$ and $\hat{Y}$ are the expected and actual prediction labels, respectively, with 1 for the favorable label and 0 for the unfavorable label. $C$ means the confidence (i.e., probability) of model prediction.  A smaller AAOD value indicates a fairer DL model.
\begin{equation}
\label{eq:AAOD}  
\small
\begin{split}
AAOD = \frac{1}{2} (|C[\hat{Y}=1|A=0,Y=0]&-C[\hat{Y}=1|A=1,Y=0]|\\ +|C[\hat{Y}=1|A=0,Y=1]&-C[\hat{Y}=1|A=1,Y=1]|).
\end{split}
\end{equation}

\item[Backward Compatibility] The backward compatibility is highly related to the occurrence of regression faults. Therefore, following existing work~\cite{you2025mitigating}, we use the negative flip rate (NFR) to measure it. As shown in Formula~\ref{eq:nfr}, NFR denotes the proportion of test inputs in the test set that trigger regression faults. The smaller the value, the more backward-compatible the DL model is.
\begin{equation}
	\label{eq:nfr}
	\mathrm{\textit{NFR}} = \frac{1}{N} \sum_{i=1}^{N} \mathcal{I}(p_{\mathcal{M}}(x_i) = y_i, p_{\mathcal{M}^\prime}(x_i) \neq y_i)
\end{equation}

\end{description}

\subsection{Implementation}
\label{sec:rqs}

We adopted the default configurations recommended in relevant papers. For the first four selected datasets in Table~\ref{table:datasets}, we preprocessed them to ensure uniform data partitioning. For datasets lacking a specific validation set, we took 10\% of the training data as the validation set (training:validation = 9:1) and split half of the original test-set data for repair following existing work~\cite{I-REPAIR}.
ImageNet is a special case. Given the high cost of training from scratch, we used the pre-trained DenseNet121 model provided by TensorFlow API. 
Please note that ImageNet does not provide the official labeled test set. Thus, following the configuration in existing work~\cite{chusupplementary,xiawindow}, we randomly select 50k samples from the training set to form a validation set. Then, we split the original validation set (which is often used as a test set in existing works~\cite{chusupplementary}) as before, with half for repair and half for testing~\cite{I-REPAIR}.

Following the recommended process from the existing study~\cite{DBLP:conf/issta/YangJS024}, we first replicated its performance on its originally provided benchmarks for each studied approach. If the results aligned with those in the papers, we adapted them to our benchmarks using the default hyperparameters reported in the original paper to ensure reliability. For approaches not originally evaluated on all our benchmarks, we followed their principles to adapt them to new datasets (e.g., using grid search for retraining-based approaches). 

We also standardized the entire evaluation process. For approaches that require extra data (e.g., validation set or a subset of the test set) for fixing, we uniformly used the dataset from the repair set to avoid any potential data leakage. During fixing, the best-performing fixed model was selected based on performance on the validation set rather than directly on the test set to avoid potential overfitting.
To avoid randomness, following existing work~\cite{you2025mitigating}, we repeated each approach on each dataset three times.
In summary, to effectively and fairly compare the experimental results of each approach, we standardized various factors such as data partitioning, evaluation process, and experiment repetition.

Our performance metrics are implemented using \textit{sklearn} and \textit{Foolbox} libraries. The hyperparameters regarding adversarial attack follow existing works~\cite{mnist_ref, mnist_utkface_ref, cifar10_ref1, cifar10_ref2, cifar10s_ref1, cifar10s_ref2}; specific epsilon values are set as follows: 0.15 for MNIST, 0.07 for UTKface, 0.031 for CIFAR10, 0.035 for CIFAR10S, and 0.007 for ImageNet. Please note that some approaches we studied are based on TensorFlow(TF), while others are based on PyTorch(PT). To ensure consistent evaluation, we used \textit{tf2onnx}~\cite{tf2onnx} and \textit{onnx2pytorch}~\cite{onnx2pytorch} to convert the TF-based models to PT. We ensured that the performance of the models remained unchanged after the conversion.
For the convenience of future research, we have implemented these models and metrics in an easy-to-use framework, which is available on our homepage.  For consistency and ease of comparison, models originally built with different library versions were migrated to a unified environment using the latest stable releases.
All experiments used Python 3.8.13, TensorFlow 2.13.1, and PyTorch 2.3.0. They were executed on an Ubuntu 20.04.3 LTS system equipped with 2.90GHz Intel(R) Xeon(R) Gold 6326 CPUs and two NVIDIA GeForce RTX 3090 GPUs.

\section{RESULTS}

\begin{table*}
\centering
\caption{Result comparison of Fixing Effectiveness of Studied Approaches}
\label{table:Result1_version3}

\scalebox{0.65}{


\begin{tabular}{cc|ccc|ccc|ccc|ccc|ccc}
\toprule
\multicolumn{2}{c}{\textbf{Dataset}}          & \multicolumn{3}{c}{MNIST}        & \multicolumn{3}{c}{UTKface}      & \multicolumn{3}{c}{CIFAR10}      & \multicolumn{3}{c}{CIFAR10S}     & \multicolumn{3}{c}{ImageNet}     \\
\midrule
\multicolumn{2}{c}{\textbf{Metric}}           & ACC   & $\Uparrow_{ACC}$ & RR    & ACC   & $\Uparrow_{ACC}$ & RR    & ACC   & $\Uparrow_{ACC}$ & RR    & ACC   & $\Uparrow_{ACC}$ & RR    & ACC   & $\Uparrow_{ACC}$ & RR    \\
\midrule
\multirow{6}{*}{Model-level}      & AP       & \cellcolor{purple!60}   99.06$_{\pm0.059}$ & \cellcolor{purple!60} 1.18 & \cellcolor{purple!60} 73.97$_{\pm2.245}$ & \cellcolor{purple!60} 90.38$_{\pm0.045}$ & \cellcolor{purple!60} 0.06 & \cellcolor{purple!60} 20.41$_{\pm0.740}$ & \cellcolor{white!58} 85.93$_{\pm0.304}$  & \cellcolor{white!58} -0.34 & \cellcolor{purple!60} 36.91$_{\pm0.958}$ & \cellcolor{white!38} 65.11$_{\pm0.073}$  & \cellcolor{white!38} -0.66 & \cellcolor{white!9} 2.62$_{\pm0.131}$    & \multicolumn{3}{c}{CORRUPTED}                                                                                    \\
                 & ET       & \cellcolor{purple!60}   99.04$_{\pm0.166}$ & \cellcolor{purple!60} 1.16 & \cellcolor{purple!60} 73.65$_{\pm1.796}$ & \multicolumn{3}{c|}{CORRUPTED}                                                                                    & \multicolumn{3}{c|}{CORRUPTED}                                                                                  & \cellcolor{white!40} 65.48$_{\pm0.282}$  & \cellcolor{white!40} -0.11 & \cellcolor{purple!65} 36.49$_{\pm0.202}$ & \multicolumn{3}{c}{CORRUPTED}                                                                                    \\
                 & FM       & \cellcolor{white!60} 98.05$_{\pm0.068}$    & \cellcolor{white!60} 0.15  & \cellcolor{white!19} 17.78$_{\pm2.245}$  & \cellcolor{purple!60} 90.49$_{\pm0.041}$ & \cellcolor{purple!60} 0.18 & \cellcolor{blue!60} 2.09$_{\pm0.427}$    & \cellcolor{white!60} 86.25$_{\pm0.025}$  & \cellcolor{white!60} 0.03  & \cellcolor{blue!60} 0.29$_{\pm0.205}$    & \cellcolor{purple!60} 65.64$_{\pm0.061}$ & \cellcolor{purple!60} 0.16 & \cellcolor{white!7} 1.71$_{\pm0.878}$    & \cellcolor{purple!60} 71.86$_{\pm0.274}$ & \cellcolor{purple!60} 0.23 & \cellcolor{purple!65} 12.96$_{\pm1.157}$ \\
                 & DR       & \cellcolor{white!39} 98.01$_{\pm0.118}$    & \cellcolor{white!39} 0.11  & \cellcolor{white!60} 22.54$_{\pm5.178}$  & \cellcolor{white!61} 90.30$_{\pm0.250}$  & \cellcolor{white!61} -0.03 & \cellcolor{white!19} 7.15$_{\pm0.653}$   & \cellcolor{purple!60} 86.36$_{\pm0.016}$ & \cellcolor{purple!60} 0.16 & \cellcolor{white!9} 2.81$_{\pm0.560}$    & \cellcolor{white!41} 65.53$_{\pm0.017}$  & \cellcolor{white!41} -0.02 & \cellcolor{blue!60} 0.13$_{\pm0.083}$    & \cellcolor{purple!65} 72.06$_{\pm0.145}$ & \cellcolor{purple!65} 0.50 & \cellcolor{white!27} 4.86$_{\pm1.946}$   \\
                 & HUDD     & \cellcolor{purple!60}   98.63$_{\pm0.066}$ & \cellcolor{purple!60} 0.75 & \cellcolor{purple!60} 50.79$_{\pm0.898}$ & \cellcolor{white!60} 89.93$_{\pm0.071}$  & \cellcolor{white!60} -0.44 & \cellcolor{white!24} 9.42$_{\pm0.000}$   & \cellcolor{purple!60} 86.47$_{\pm0.465}$ & \cellcolor{purple!60} 0.29 & \cellcolor{purple!60} 36.48$_{\pm0.247}$ & \cellcolor{purple!65} 68.82$_{\pm0.375}$ & \cellcolor{purple!60} 5.00 & \cellcolor{purple!60} 28.64$_{\pm0.792}$ & \cellcolor{purple!60} 71.71$_{\pm0.014}$ & \cellcolor{purple!60} 0.01 & \cellcolor{white!8} 0.68$_{\pm0.058}$    \\
                 & HR       & \cellcolor{white!39} 98.00$_{\pm0.465}$    & \cellcolor{white!39} 0.11  & \cellcolor{white!60} 27.30$_{\pm4.283}$  & \cellcolor{white!60} 90.32$_{\pm0.147}$  & \cellcolor{white!60} -0.01 & \cellcolor{white!15} 5.23$_{\pm0.740}$   & \cellcolor{purple!60} 86.49$_{\pm0.170}$ & \cellcolor{purple!60} 0.32 & \cellcolor{white!60} 6.92$_{\pm1.368}$   & \cellcolor{white!38} 65.10$_{\pm0.460}$  & \cellcolor{white!38} -0.67 & \cellcolor{white!16} 6.78$_{\pm3.841}$   & \multicolumn{3}{c}{CORRUPTED}                                                                                    \\  \midrule
\multirow{2}{*}{Layer-level}     & 3M       & \cellcolor{white!36} 97.90$_{\pm0.000}$    & \cellcolor{white!36} 0.00  & \cellcolor{blue!60} 0.00$_{\pm0.000}$    & \cellcolor{blue!60} 88.77$_{\pm0.000}$   & \cellcolor{blue!60} -1.73  & \cellcolor{purple!60} 17.27$_{\pm0.000}$ & \cellcolor{white!61} 86.22$_{\pm0.000}$  & \cellcolor{white!61} 0.00  & \cellcolor{blue!60} 0.00$_{\pm0.000}$    & \cellcolor{purple!60} 65.58$_{\pm0.000}$ & \cellcolor{purple!60} 0.06 & \cellcolor{white!7} 1.25$_{\pm0.000}$    & \multicolumn{3}{c}{FAILURE}                                                                                       \\
                 & RA       & \cellcolor{white!33} 97.96$_{\pm0.000}$    & \cellcolor{white!33} -0.14 & \cellcolor{blue!60} 6.67$_{\pm0.000}$    & \multicolumn{3}{c}{UNADAPTABLE}                                                                                    & \cellcolor{blue!60} 85.76$_{\pm0.000}$   & \cellcolor{blue!60} -0.53  & \cellcolor{white!11} 4.21$_{\pm0.000}$   & \multicolumn{3}{c|}{UNADAPTABLE}                                                                                    & \multicolumn{3}{c}{UNADAPTABLE}                                                                                    \\ \midrule
\multirow{8}{*}{Neuron-level}     & AR       & \cellcolor{white!33} 97.75$_{\pm0.210}$    & \cellcolor{white!33} -0.15 & \cellcolor{white!15} 12.70$_{\pm4.687}$  & \cellcolor{purple!60} 90.34$_{\pm0.066}$ & \cellcolor{purple!60} 0.01 & \cellcolor{blue!60} 1.92$_{\pm1.305}$    & \cellcolor{white!61} 86.21$_{\pm0.164}$  & \cellcolor{white!61} -0.01 & \cellcolor{white!13} 5.42$_{\pm2.988}$   & \cellcolor{white!40} 65.39$_{\pm0.190}$  & \cellcolor{white!40} -0.23 & \cellcolor{white!12} 4.29$_{\pm1.128}$   & \cellcolor{white!55} 71.68$_{\pm0.016}$  & \cellcolor{white!55} -0.03 & \cellcolor{white!6} 0.34$_{\pm0.087}$    \\
                 & CARE     & \cellcolor{blue!60} 97.59$_{\pm0.093}$     & \cellcolor{blue!60} -0.32  & \cellcolor{white!16} 13.97$_{\pm3.143}$  & \cellcolor{blue!60} 89.71$_{\pm0.477}$   & \cellcolor{blue!60} -0.69  & \cellcolor{white!60} 12.39$_{\pm6.256}$  & \cellcolor{white!59} 86.03$_{\pm0.019}$  & \cellcolor{white!59} -0.22 & \cellcolor{white!6} 1.16$_{\pm0.119}$    & \cellcolor{white!41} 65.54$_{\pm0.016}$  & \cellcolor{white!41} 0.00  & \cellcolor{blue!60} 0.11$_{\pm0.027}$    & \cellcolor{white!56} 71.70$_{\pm0.000}$  & \cellcolor{white!56} 0.00  & \cellcolor{blue!60} 0.00$_{\pm0.000}$    \\
                 & NR       & \cellcolor{blue!60} 96.73$_{\pm0.996}$     & \cellcolor{blue!60} -1.20  & \cellcolor{white!12} 9.84$_{\pm2.500}$   & \cellcolor{white!54} 89.96$_{\pm0.232}$  & \cellcolor{white!54} -0.41 & \cellcolor{white!17} 6.11$_{\pm1.075}$   & \cellcolor{white!56} 85.80$_{\pm0.325}$  & \cellcolor{white!56} -0.49 & \cellcolor{white!60} 6.87$_{\pm0.694}$   & \multicolumn{3}{c|}{FAILURE}                                                                                       & \multicolumn{3}{c}{UNADAPTABLE}                                                                                    \\
                 & IR       & \cellcolor{white!60} 98.06$_{\pm0.016}$    & \cellcolor{white!60} 0.16  & \cellcolor{white!17} 15.56$_{\pm1.796}$  & \cellcolor{blue!60} 87.43$_{\pm0.391}$   & \cellcolor{blue!60} -3.21  & \cellcolor{purple!60} 29.48$_{\pm0.890}$ & \cellcolor{blue!60} 81.54$_{\pm0.490}$   & \cellcolor{blue!60} -5.43  & \cellcolor{purple!60} 19.59$_{\pm0.721}$ & \cellcolor{blue!60} 64.87$_{\pm0.175}$   & \cellcolor{blue!60} -1.02  & \cellcolor{white!18} 8.43$_{\pm0.302}$   & \cellcolor{blue!60} 70.27$_{\pm0.463}$   & \cellcolor{blue!60} -1.99  & \cellcolor{purple!60} 6.71$_{\pm1.128}$  \\
                 & DTR      & \multicolumn{3}{c|}{UNADAPTABLE}                                                                                      & \multicolumn{3}{c|}{UNADAPTABLE}                                                                                    & \cellcolor{blue!60} 85.63$_{\pm0.239}$    & \cellcolor{blue!60} -0.68  & \cellcolor{white!7} 1.40$_{\pm0.416}$    & \cellcolor{blue!60} 64.13$_{\pm0.377}$   & \cellcolor{blue!60} -2.16  & \cellcolor{white!12} 4.51$_{\pm0.000}$   & \cellcolor{blue!60} 70.43$_{\pm1.0300}$   & \cellcolor{blue!60} -1.78  & \cellcolor{blue!60} 0.20$_{\pm0.093}$    \\
                 & GM       & \cellcolor{white!38} 97.95$_{\pm0.053}$    & \cellcolor{white!38} 0.05  & \cellcolor{blue!60} 3.49$_{\pm4.283}$    & \cellcolor{white!60} 90.33$_{\pm0.000}$  & \cellcolor{white!60} 0.00  & \cellcolor{blue!60} 0.00$_{\pm0.000}$    & \cellcolor{white!61} 86.22$_{\pm0.000}$  & \cellcolor{white!61} 0.00  & \cellcolor{blue!60} 0.00$_{\pm0.000}$    & \cellcolor{white!41} 65.54$_{\pm0.000}$  & \cellcolor{white!41} 0.00  & \cellcolor{blue!60} 0.00$_{\pm0.000}$    & \cellcolor{white!56} 71.70$_{\pm0.000}$  & \cellcolor{white!56} 0.00  & \cellcolor{blue!60} 0.00$_{\pm0.000}$    \\
                 & VERE     & \multicolumn{3}{c|}{UNADAPTABLE}                                                                                      & \multicolumn{3}{c|}{UNADAPTABLE}                                                                                    & \cellcolor{white!60} 86.23$_{\pm0.019}$  & \cellcolor{white!60} 0.01  & \cellcolor{white!5} 0.53$_{\pm0.068}$    & \multicolumn{3}{c|}{UNADAPTABLE}       &                                                                              \multicolumn{3}{c}{UNADAPTABLE}                                                                                    \\
                 & IN       & \cellcolor{blue!60} 96.59$_{\pm0.009}$     & \cellcolor{blue!60} -1.34  & \cellcolor{white!22} 21.90$_{\pm0.000}$  & \cellcolor{white!58} 90.16$_{\pm0.118}$  & \cellcolor{white!58} -0.18 & \cellcolor{white!60} 12.73$_{\pm2.019}$  & \cellcolor{white!58} 86.07$_{\pm0.094}$  & \cellcolor{white!58} -0.27 & \cellcolor{white!60} 8.85$_{\pm6.261}$   & \cellcolor{blue!60} 60.53$_{\pm0.065}$   & \cellcolor{blue!60} -7.64  & \cellcolor{purple!60} 13.25$_{\pm0.178}$ & \cellcolor{blue!60} 69.55$_{\pm0.000}$   & \cellcolor{blue!60} -3.00  & \cellcolor{purple!60} 8.93$_{\pm0.020}$ \\
                              \bottomrule
\end{tabular}
}
\begin{tablenotes}
\scriptsize
\item[*] The $\Uparrow_{ACC}$ represents the relative improvement in the model's accuracy after the fixing process.
\item[*] We use color coding to highlight metric values: purple denotes top-3 rankings, and blue indicates bottom 3.
\item[*] The values are presented as a$\pm$b, where a denotes the mean and b the standard deviation across three repetitions of the experiment.
\end{tablenotes}


\end{table*}

\begin{table}
\centering
\caption{T Test Analysis of Studied Approaches}
\label{table:groups}

\scalebox{0.75}{


\begin{tabular}{cc|c|c|c|c|c|c}
\toprule
\multicolumn{2}{c|}{\textbf{Approach}} & \textbf{ACC} & \textbf{RR} & \textbf{ASR} & \textbf{AAOD} & \textbf{NFR} & \textbf{TIME} \\
\midrule
\multirow{6}{*}{Model-level}   & AP   & ab           & a           & a            & a             & ef           & g             \\
                               & ET   & b            & a           & a            & e             & g            & g             \\
                               & FM   & ab           & c           & c            & g             & cd           & cd            \\
                               & DR   & ab           & c           & c            & a             & d            & e             \\
                               & HUDD   & a            & a           & g            & a             & e            & e             \\
                               & HR   & c            & b           & e            & b             & dfg          & d             \\
                               \midrule
\multirow{2}{*}{Layer-level}   & 3M   & cef          & c           & bc           & g             & d            & a             \\
                               & RA   & g            & d           & ef           & -             & c            & a             \\
                               \midrule
\multirow{8}{*}{Neuron-level}  & AR   & e            & c           & f            & f             & d            & c             \\
                               & CARE & f            & c           & b            & b             & d            & de            \\
                               & NR   & f            & c           & b            & d             & e            & e             \\
                               & IR   & gh           & b           & b            & f             & g            & c             \\
                               & DTR  & h            & e           & c            & c             & f            & d             \\
                               & GM   & c            & f           & e            & f             & a            & f             \\
                               & VERE & d            & g           & d            & -             & b            & b             \\
                               & IN   & gh           & b           & bce          & a             & g            & c  \\
                               \bottomrule
\end{tabular}
}
\begin{tablenotes}
\scriptsize
\item[*] VERE and RA cannot apply to CIFAR10S and UTKFace thus do not have AAOD
\end{tablenotes}


\end{table}

\begin{table*}
\centering
\caption{Result Comparison of Adversarial Robustness, Fairness and Backward Compatibility of Studied Approaches}
\label{table:Result2_version3}
\scalebox{0.57}{


\begin{tabular}{cc|ccc|ccccc|ccc|ccccc|ccc}
\toprule
\multicolumn{2}{c|}{\textbf{Dataset}}          & \multicolumn{3}{c|}{MNIST}       & \multicolumn{5}{c|}{UTKface}                                  & \multicolumn{3}{c|}{CIFAR10}     & \multicolumn{5}{c|}{CIFAR10S}                                  & \multicolumn{3}{c}{ImageNet}    \\
\midrule
\multicolumn{2}{c|}{\textbf{Metric}}           & ASR   & $\Uparrow_{ASR}$ & NFR  & ASR   & $\Uparrow_{ASR}$ & AAOD   & $\Uparrow_{AAOD}$ & NFR  & ASR   & $\Uparrow_{ASR}$ & NFR  & ASR   & $\Uparrow_{ASR}$ & AAOD   & $\Uparrow_{AAOD}$ & NFR   & ASR   & $\Uparrow_{ASR}$ & NFR  \\
\midrule
\multirow{6}{*}{Model-level}  & AP   & \cellcolor{purple!60}23.17$_{\pm0.957}$ & \cellcolor{purple!60}69.87            & \cellcolor{white!51}0.39$_{\pm0.041}$ & \cellcolor{blue!60}91.53$_{\pm0.957}$ & \cellcolor{blue!60}-6.80            & \cellcolor{purple!60}0.0140$_{\pm0.001}$ & \cellcolor{purple!60}30.53             & \cellcolor{blue!60}1.92$_{\pm0.041}$ & \cellcolor{purple!60}20.90$_{\pm0.772}$ & \cellcolor{purple!60}67.85            & \cellcolor{blue!60}5.37$_{\pm0.180}$ & \cellcolor{white!18}76.20$_{\pm0.206}$ & \cellcolor{white!18}-1.09            & \cellcolor{white!35}0.3164$_{\pm0.001}$ & \cellcolor{white!35}0.77              & \cellcolor{white!21}1.33$_{\pm0.082}$  & \multicolumn{3}{c}{CORRUPTED}   \\
                              & ET   & \cellcolor{purple!60}15.93$_{\pm9.028}$ & \cellcolor{purple!60}79.76            & \cellcolor{white!51}0.41$_{\pm0.131}$ & \multicolumn{5}{c|}{CORRUPTED}                                & \multicolumn{3}{c|}{CORRUPTED}   & \cellcolor{purple!60}66.90$_{\pm9.328}$ & \cellcolor{purple!60}23.98            & \cellcolor{white!37}0.3146$_{\pm0.003}$ & \cellcolor{white!37}1.33              & \cellcolor{blue!60}12.45$_{\pm0.172}$ & \multicolumn{3}{c}{CORRUPTED}   \\
                              & FM   & \cellcolor{white!18}82.33$_{\pm0.579}$ & \cellcolor{white!18}-0.90            & \cellcolor{white!56}0.25$_{\pm0.084}$ & \cellcolor{white!31}84.97$_{\pm0.896}$ & \cellcolor{white!31}1.09             & \cellcolor{blue!60}0.0230$_{\pm0.001}$ & \cellcolor{blue!60}-14.03            & \cellcolor{purple!60}0.05$_{\pm0.041}$ & \cellcolor{white!10}65.37$_{\pm0.124}$ & \cellcolor{white!10}-0.26            & \cellcolor{purple!60}0.01$_{\pm0.009}$ & \cellcolor{white!20}80.53$_{\pm0.170}$ & \cellcolor{white!20}-0.17            & \cellcolor{blue!60}0.3199$_{\pm0.000}$ & \cellcolor{blue!60}-0.35             & \cellcolor{white!64}0.49$_{\pm0.266}$  & \cellcolor{purple!60}82.30$_{\pm0.283}$ & \cellcolor{purple!60}3.18             & \cellcolor{blue!60}3.51$_{\pm0.061}$ \\
                              & DR   & \cellcolor{white!16}84.77$_{\pm1.097}$ & \cellcolor{white!16}-3.38            & \cellcolor{white!52}0.37$_{\pm0.096}$ & \cellcolor{purple!60}81.50$_{\pm0.497}$ & \cellcolor{purple!60}3.78            & \cellcolor{white!49}0.0173$_{\pm0.003}$ & \cellcolor{white!49}13.70             & \cellcolor{white!57}0.73$_{\pm0.310}$ & \cellcolor{white!9}68.57$_{\pm0.450}$ & \cellcolor{white!9}-2.19            & \cellcolor{white!63}0.23$_{\pm0.0618}$ & \cellcolor{white!21}80.63$_{\pm0.206}$ & \cellcolor{white!21}0.33             & \cellcolor{white!32}0.3187$_{\pm0.000}$ & \cellcolor{white!32}0.03              & \cellcolor{purple!60}0.07$_{\pm0.237}$  & \cellcolor{purple!60}81.47$_{\pm0.368}$ & \cellcolor{purple!60}0.89             & \cellcolor{white!53}1.02$_{\pm0.408}$ \\
                              & HUDD & \cellcolor{blue!60}86.53$_{\pm1.970}$ & \cellcolor{blue!60}-7.36            & \cellcolor{white!53}0.33$_{\pm0.077}$ & \cellcolor{blue!60}76.80$_{\pm0.779}$ & \cellcolor{blue!60}-4.07            & \cellcolor{purple!60}0.0154$_{\pm0.002}$ & \cellcolor{purple!60}23.76             & \cellcolor{white!51}1.32$_{\pm0.072}$ & \cellcolor{blue!60}69.63$_{\pm1.893}$ & \cellcolor{blue!60}-7.79            & \cellcolor{blue!60}4.77$_{\pm0.433}$ & \cellcolor{blue!60}78.23$_{\pm1.597}$ & \cellcolor{blue!60}-8.65            & \cellcolor{blue!60}0.3411$_{\pm0.002}$ & \cellcolor{blue!60}-6.99             & \cellcolor{white!53}6.59$_{\pm0.107}$  & \cellcolor{white!18}80.33$_{\pm0.047}$ & \cellcolor{white!18}-0.04            & \cellcolor{white!62}0.18$_{\pm0.012}$ \\
                              & HR   & \cellcolor{white!18}79.10$_{\pm1.499}$ & \cellcolor{white!18}0.25            & \cellcolor{white!47}0.53$_{\pm0.172}$ & \cellcolor{white!25}84.77$_{\pm0.386}$ & \cellcolor{white!25}-0.68             & \cellcolor{white!41}0.0195$_{\pm0.001}$ & \cellcolor{white!41}3.47             & \cellcolor{white!59}0.52$_{\pm0.170}$ & \cellcolor{white!9}64.47$_{\pm0.881}$ & \cellcolor{white!9}-2.01            & \cellcolor{white!53}1.35$_{\pm0.531}$ & \cellcolor{white!17}78.10$_{\pm0.990}$ & \cellcolor{white!17}-1.83            & \cellcolor{purple!60}0.3123$_{\pm0.008}$ & \cellcolor{purple!60}2.04             & \cellcolor{blue!60}34.27$_{\pm1.485}$ & \multicolumn{3}{c}{CORRUPTED}   \\
\midrule
\multirow{2}{*}{Layer-level}  & 3M   & \cellcolor{white!16}81.93$_{\pm1.050}$ & \cellcolor{white!16}-3.71            & \cellcolor{purple!60}0.00$_{\pm0.000}$ & \cellcolor{purple!60}72.77$_{\pm0.170}$ & \cellcolor{purple!60}11.15            & \cellcolor{blue!60}0.0291$_{\pm0.000}$ & \cellcolor{blue!60}-44.06            & \cellcolor{blue!60}3.24$_{\pm0.000}$ & \cellcolor{blue!60}73.53$_{\pm0.896}$ & \cellcolor{blue!60}-3.42            & \cellcolor{purple!60}0.00$_{\pm0.000}$ & \cellcolor{white!16}81.27$_{\pm0.759}$ & \cellcolor{white!16}-2.36            & \cellcolor{blue!60}0.3189$_{\pm0.000}$ & \cellcolor{blue!60}-0.03             & \cellcolor{white!64}0.39$_{\pm0.000}$  & \multicolumn{3}{c}{FAILURE}     \\
                              & RA   & \cellcolor{blue!60}89.20$_{\pm0.000}$ & \cellcolor{blue!60}-10.26           & \cellcolor{purple!60}0.14$_{\pm0.000}$ & \multicolumn{5}{c|}{UNADAPTABLE}                              & \cellcolor{white!10}75.20 & \cellcolor{white!10}-0.67            & \cellcolor{white!64}0.12 & \multicolumn{5}{c|}{UNADAPTABLE}                               & \multicolumn{3}{c}{UNADAPTABLE} \\

\midrule
\multirow{8}{*}{Neuron-level} & AR   & \cellcolor{white!16}85.73$_{\pm3.087}$ & \cellcolor{white!16}-3.29            & \cellcolor{white!51}0.41$_{\pm0.118}$ & \cellcolor{blue!60}82.60$_{\pm1.0678}$ & \cellcolor{blue!60}-1.47            & \cellcolor{blue!60}0.0225$_{\pm0.003}$ & \cellcolor{blue!60}-11.39            & \cellcolor{purple!60}0.17$_{\pm0.063}$ & \cellcolor{blue!60}67.80$_{\pm1.846}$ & \cellcolor{blue!60}-4.95            & \cellcolor{white!58}0.75$_{\pm0.474}$ & \cellcolor{white!16}75.10$_{\pm0.143}$ & \cellcolor{white!16}-2.46            & \cellcolor{white!33}0.3180$_{\pm0.001}$ & \cellcolor{white!33}0.24              & \cellcolor{white!62}1.63$_{\pm0.325}$  & \cellcolor{blue!60}84.60$_{\pm0.638}$ & \cellcolor{blue!60}-1.08            & \cellcolor{purple!60}0.11$_{\pm0.036}$ \\
                              & CARE & \cellcolor{purple!60}81.00$_{\pm0.668}$ & \cellcolor{purple!60}0.61             & \cellcolor{blue!60}0.61$_{\pm0.151}$ & \cellcolor{white!31}84.60$_{\pm0.779}$ & \cellcolor{white!31}1.05             & \cellcolor{white!42}0.0194$_{\pm0.001}$ & \cellcolor{white!42}4.13              & \cellcolor{white!45}1.82$_{\pm1.003}$ & \cellcolor{white!11}66.90$_{\pm0.849}$ & \cellcolor{white!11}0.89             & \cellcolor{white!62}0.35$_{\pm0.009}$ & \cellcolor{purple!60}78.67$_{\pm0.047}$ & \cellcolor{purple!60}0.42             & \cellcolor{white!33}0.3180$_{\pm0.001}$ & \cellcolor{white!33}0.26              & \cellcolor{purple!60}0.04$_{\pm0.025}$  & \cellcolor{white!12}84.70$_{\pm0.779}$ & \cellcolor{white!12}-0.47            & \cellcolor{purple!60}0.00$_{\pm0.000}$ \\
                              & NR   & \cellcolor{white!18}79.53$_{\pm0.685}$ & \cellcolor{white!18}-0.42            & \cellcolor{blue!60}1.38$_{\pm0.947}$ & \cellcolor{white!33}85.07$_{\pm0.776}$ & \cellcolor{white!33}1.77             & \cellcolor{white!40}0.0199$_{\pm0.001}$ & \cellcolor{white!40}1.49              & \cellcolor{white!54}0.96$_{\pm0.338}$ & \cellcolor{white!14}59.70$_{\pm1.374}$ & \cellcolor{white!14}4.17             & \cellcolor{white!53}1.37$_{\pm0.420}$ & \multicolumn{5}{c|}{FAILURE}                                   & \multicolumn{3}{c}{UNADAPTABLE} \\
                              & IR   & \cellcolor{white!16}83.77$_{\pm0.403}$ & \cellcolor{white!16}-2.91            & \cellcolor{white!57}0.23$_{\pm0.093}$ & \cellcolor{purple!60}77.93$_{\pm0.287}$ & \cellcolor{purple!60}6.00             & \cellcolor{white!32}0.0220$_{\pm0.003}$ & \cellcolor{white!32}-8.91             & \cellcolor{blue!60}5.75$_{\pm0.442}$ & \cellcolor{purple!60}51.17$_{\pm1.204}$ & \cellcolor{purple!60}21.76            & \cellcolor{blue!60}7.36$_{\pm0.508}$ & \cellcolor{purple!60}76.63$_{\pm0.249}$ & \cellcolor{purple!60}0.87             & \cellcolor{white!36}0.3159$_{\pm0.001}$ & \cellcolor{white!36}0.92              & \cellcolor{white!58}3.60$_{\pm0.178}$  & \cellcolor{blue!60}83.60$_{\pm0.374}$ & \cellcolor{blue!60}-0.60            & \cellcolor{blue!60}3.33$_{\pm0.781}$ \\
                              & DTR  & \multicolumn{3}{c|}{UNADAPTABLE} & \multicolumn{5}{c|}{UNADAPTABLE}                              & \cellcolor{white!14}59.73$_{\pm0.643}$ & \cellcolor{white!14}4.73             & \cellcolor{white!58}0.75$_{\pm0.289}$ & \cellcolor{blue!60}85.20$_{\pm1.657}$ & \cellcolor{blue!60}-4.80            & \cellcolor{purple!60}0.3108$_{\pm0.001}$ & \cellcolor{purple!60}2.50              & \cellcolor{white!59}2.97$_{\pm0.486}$  & \cellcolor{blue!60}83.27$_{\pm0.838}$ & \cellcolor{blue!60}-1.05            & \cellcolor{white!50}1.32$_{\pm1.049}$ \\
                              & GM   & \cellcolor{white!16}81.10$_{\pm0.653}$ & \cellcolor{white!16}-2.92            & \cellcolor{purple!60}0.03$_{\pm0.038}$ & \cellcolor{white!27}83.50$_{\pm0.000}$ & \cellcolor{white!27}0.00             & \cellcolor{white!39}0.0202$_{\pm0.000}$ & \cellcolor{white!39}0.00              & \cellcolor{purple!60}0.00$_{\pm0.000}$ & \cellcolor{white!10}67.47$_{\pm0.094}$ & \cellcolor{white!10}-0.40            & \cellcolor{purple!60}0.00$_{\pm0.000}$ & \cellcolor{white!15}75.27$_{\pm1.934}$ & \cellcolor{white!15}-2.75            & \cellcolor{white!32}0.3188$_{\pm0.000}$ & \cellcolor{white!32}0.00              & \cellcolor{purple!60}0.00$_{\pm0.000}$  & \cellcolor{white!23}83.60$_{\pm0.374}$ & \cellcolor{white!23}0.36             & \cellcolor{purple!60}0.00$_{\pm0.000}$ \\
                              & VERE & \multicolumn{3}{c|}{UNADAPTABLE} & \multicolumn{5}{c|}{UNADAPTABLE}                              & \cellcolor{white!10}72.90$_{\pm0.163}$ & \cellcolor{white!10}-0.83            & \cellcolor{white!64}0.06$_{\pm0.028}$ & \multicolumn{5}{c|}{UNADAPTABLE}                               & \multicolumn{3}{c}{UNADAPTABLE} \\
                              & IN   & \cellcolor{blue!60}99.70$_{\pm0.000}$ & \cellcolor{blue!60}-23.24           & \cellcolor{blue!60}1.77$_{\pm0.009}$ & \cellcolor{white!27}86.43$_{\pm0.206}$ & \cellcolor{white!27}0.08             & \cellcolor{purple!60}0.0134$_{\pm0.001}$ & \cellcolor{purple!60}33.66             & \cellcolor{white!50}1.43$_{\pm0.186}$ & \cellcolor{purple!60}58.03$_{\pm5.280}$ & \cellcolor{purple!60}11.40            & \cellcolor{white!53}1.35$_{\pm0.957}$ & \cellcolor{blue!60}79.93$_{\pm0.189}$ & \cellcolor{blue!60}-4.62            & \cellcolor{purple!65}0.2926$_{\pm0.001}$ & \cellcolor{purple!65}8.23              & \cellcolor{blue!60}9.60$_{\pm0.013}$  & \cellcolor{purple!60}81.00$_{\pm0.327}$ & \cellcolor{purple!60}3.69             & \cellcolor{blue!60}5.30$_{\pm0.008}$ \\
                              \bottomrule
\end{tabular}
}

\begin{tablenotes}
\scriptsize
\item[*] The $\Uparrow_{AAOD}$ and $\Uparrow_{ASR}$ represent the relative improvement in the model's AAOD and ASR after the fixing process.
\item[*] We use color coding to highlight metric values: purple denotes top-3 rankings, and blue indicates bottom 3.
\item[*] The values are presented as a$\pm$b, where a denotes the mean and b the standard deviation across three repetitions of the experiment.

\end{tablenotes}


\end{table*}

\subsection{RQ1: Applicability}

From Table~\ref{table:Result1_version3}, we observe that not all approaches can be successfully adapted to other models or datasets. We classify their inapplicability into three categories:
1) \textbf{CORRUPTED}: Approaches can be implemented on the models/structures, but they degrade the model's accuracy by over 10\%.
2) \textbf{FAILURE}: During the adaptation of these approaches, certain bugs arise. For example, 3M fails to work on ImageNet due to the exorbitant calculation costs of constraint solving on DenseNet121, resulting in out-of-memory issues. NR fails on CIFAR10S as the version-to-version difference is insufficient to identify neurons responsible for faults.
3) \textbf{UNADAPTABLE}: Approaches are incompatible with model structures or datasets. For example, RA requires the models to have at least three consecutive fully-connected layers. VERE currently supports fixing fully-connected layers with the same input and output dimensions. For NR, we lack historical versions for ImageNet-DenseNet121. DTR is required to work on datasets where class importance matters in the given tasks. Based on its GitHub repository~\cite{DTRgithub}, DTR is applied to CIFAR10 by prioritizing \textit{vehicle-related classes}, an assumption also mentioned in prior work~\cite{DBLP:conf/issre/DuranZAI21}. In our study, we extended this setting to CIFAR10S and ImageNet and then applied DTR to these three datasets with class importance settings. DTR was not adapted to the other two datasets (i.e., MNIST and UTKFace), which lack class importance settings, following the insight of DTR.

In summary, we assessed the applicability of different-level fixing approaches across our datasets. Among model-level fixing approaches, three out of six are suitable for all datasets. For neuron-level fixing approaches, five out of eight exhibit broad applicability, while none of the layer-level fixing approaches are applicable to all datasets. This shows that neuron-level and model-level fixing approaches have relatively greater scalability.
Specifically, among model-level approaches, FM, DR, and HUDD are more scalable, primarily because they modify training data slightly via augmentation or selection. In contrast, drastic modifications to objective functions (ET) and weights (AP) carry a higher risk of degrading overall performance.
Regarding neuron-level approaches, AR\&IR (gradient-based), GM (spectrum-based), CARE (causality-based), and IN (feature-based) are more scalable. They require neither external information (e.g., historical versions) nor additional requirements (e.g., verification) to guide fixing.

\finding{In general, neuron and model-level fixing approaches have stronger applicability than layer-level approaches. Among model-level approaches, FM, DR, and HUDD achieve better scalability through minimal data modification. Among neuron-level approaches, AR, IR, GM, CARE, and IN avoid using external guidance for fault localization and fixing.}

\subsection{RQ2: Effectiveness}

As described in Sections~\ref{sec:studied_approaches} and \ref{sec:exmperiment}, we systematically studied 16 DL fixing approaches across five different datasets. The experimental results are presented in Table~\ref{table:Result1_version3}.
To analyze performance differences, we group approaches using the T-Test~\cite{boneau1960effects} (significance level is 0.05) with Benjamini-Hochberg correction~\cite{benjamini1995controlling} to identify statistically significant differences following existing work~\cite{DBLP:journals/tosem/ZhouCH22,zheng2025identifying}. Approaches are clustered into groups when no significant differences exist between them, and alphabet labels (\textit{a}, \textit{b}, \textit{c}) are assigned such that earlier labels denote superior performance (\textit{a} > \textit{b} > \textit{c}). Please note that some approaches may have more than one alphabet label. Table~\ref{table:groups} presents the results.

Regarding overall accuracy improvement, model-level approaches consistently outperform others across datasets. On average, model-level approaches yield a 0.17\% relative accuracy gain, compared to -0.50\% for layer-level and -1.01\% for neuron-level approaches. Specifically, 87\% (13/15) of top-3 performance results fall into model-level approaches. This advantage stems from model-level approaches optimizing parameters globally, preserving inter-neuron synergy to systematically correct faulty behaviors via data distribution learning~\cite{HybridRepair}. In contrast, individual neuron modifications may disrupt synergy and induce overfitting to local features. As shown in Table~\ref{table:Result1_version3}, 87\% (13/15) of bottom-3 performance results fall in neuron-level approaches. Within model-level approaches: HUDD achieves four top-3 accuracy improvements, particularly on CIFAR10S (5\% gain), likely due to its strategy of selecting representative faulty samples for repair. FM maintains stable positive gains across all datasets by modestly augmenting data distributions with lightweight finetuning. For neuron-level approaches, performance is generally poor. This is evident in IR, which causes severe accuracy drops on four datasets. The primary reason is its aggressive backward-propagation adjustments, exacerbating overfitting. However, VERE and GM avoid degradation on all datasets by adopting conservative fixing strategies: VERE uses verification-guided constraints, while GM strictly limits mutation degree. We also note that although layer-level approaches perform poorly in fault fixing, their fault-fixing effectiveness is more stable due to a lower standard deviation than other training-based approaches (e.g., ET). This is primarily because they adopt techniques such as verification and linear programming solving to avoid significant changes in the model weights.
Overall, maintaining accuracy during fixing is challenging. Only three approaches, i.e., FM, GM, and VERE, avoid degradation across all datasets.

\finding{In terms of overall accuracy, model-level approaches outperform layer-level, which in turn outperform neuron-level. 
HUDD (model-level) achieves relatively better accuracy improvement, while IR (neuron-level) performs relatively worse. 
Approaches with conservative changes (FM from model-level and GM\&VERE from neuron-level) stably preserve accuracy.}

Regarding error correction capability (i.e., RR), model-level approaches demonstrate superior performance. Most (3/6) belong to group $\textit{a}$ in Table~\ref{table:groups}, achieving an average RR of 14.82\% compared to 7.32\% for neuron-level and 5.99\% for layer-level approaches. Model-level superiority stems from global weight adjustments that better address fault-inducing data. Layer-level approaches underperform due to their reliance on constraint-solving/verification and reluctance to modify model weights. Notably, for model-level, retraining-based approaches (AP, ET, HUDD) outperform finetuning-based ones (DR, FM, HR), e.g., AP outperforms FM on MNIST with an average improvement of 316.03\%; this is primarily due to their more radical modifications (e.g., directly changing loss functions, weights, or data distributions). Similarly, for neuron-level, IR (focal-tuning), IN (feature alignment) outperform search-based ones, e.g., IR outperforms NR on CIFAR10 with an average improvement of 185.15\%; primarily because they drastically modify neurons to align with ground truth. The approach with the worst performance is GM. The RRs of GM are zero on four datasets, and it does not fix anything on these models.
These results suggest that approaches with radical model modification have stronger error correction capabilities. 

Correcting model behavior while maintaining overall accuracy remains challenging. Analysis of all cases reveals current approaches fix over 5\% of faults (RR) in only 11 out of 63 cases without accuracy loss, with 10 of these being model-level approaches. These results highlight the critical need for developers to balance error correction capability and accuracy preservation.

\finding{
Regarding error correction capability, model-level approaches outperform neuron-level ones, which in turn surpass layer-level ones. 
For model-level, retraining-based approaches (e.g., AP, ET) generally outperform finetuning-based ones (e.g., DR, FM). 
For neuron-level, training-based approaches (e.g., IR, IN) generally outperform search-based ones (e.g., GM, AR). 
Approaches incorporating radical modifications tend to exhibit stronger error correction capabilities.
}

We observe an interesting phenomenon: the overall effectiveness of fixing approaches declines as model complexity increases. Take AP (weight-adjustment-based approach) as an example. When applied to the MNIST dataset with LeNet5, AP achieves an RR of over 70\% and an accuracy improvement of 1.18\%. However, on the CIFAR10S dataset with ResNet18, both metrics deteriorate. The RR drops to merely 2.62\%, and the overall accuracy slightly decreases by 0.66\%. On the ImageNet dataset with DenseNet121, AP fails to perform the fixing task, as its accuracy plummets by more than 10\%, which is unacceptable. This finding implies that current fixing approaches are more effective for small-scale models, and we need more effective fixing approaches for large-scale models. 
Please note that the finetuning/retraining-based approaches (e.g., FM, DR, HUDD) from the model level have relatively higher RR without sacrificing accuracy on ImageNet, while the search-based approaches from the neuron level (e.g., AR) achieve a minimal repair rate (less than 0.34\%). This indicates that finetuning/retraining methods have the potential to work on large-scale models.

\finding{
Current fixing approaches suit small-scale models better. 
More approaches for large-scale models are needed.
}

\subsection{RQ3: Impacts on Other Properties}

Regarding robustness measured by ASR, the top-performing approaches from Table~\ref{table:groups} group $\textit{a}$ are AP (weight-adjustment-based) and ET (objective function-based), achieving ASR of nearly 70\% on MNIST. This is likely due to their effective exploitation of white-box information. Notably, ET (objective function-based, model-level) emerges as the approach with relatively higher RR and greater robustness gains, as it avoids robustness degradation across all datasets. This superiority likely stems from ET's gradient descent editor mechanism strategy, which facilitates robust training. However, due to the drastic changes resulting from objective function adjustments and training processes, ET also exhibits the highest standard deviation.
The worst performance overall is HUDD (counterexample based). HUDD consistently reduces robustness across all scenarios, possibly due to robust overfitting~\cite{DBLP:conf/nips/RebuffiGCSWM21} induced by limited failing samples during repair.
Among neuron-level approaches, IR demonstrates relatively good robustness improvement (achieving top-3 on three datasets). This is mainly because it modifies only parameters significantly impacting misclassified inputs, thereby preserving robustness. In contrast, AR performs worst in neuron-level approaches, ranking bottom-3 on three datasets. This might be due to AR's reliance on certain faulty inputs to guide fixing, which could induce robust overfitting~\cite{DBLP:conf/nips/RebuffiGCSWM21} during optimization.

\finding{
Regarding ASR, model-level approaches are more likely to preserve robustness compared to neuron-level and layer-level approaches. Specifically, AP and ET (model-level) demonstrate superior robustness improvements due to their effective utilization of white-box information. ET is the most stable approach that preserves robustness.
}

In terms of fairness, current approaches generally preserve fairness. Among model-level approaches, three of them (AP, DR, HUDD) belong to group $\textit{a}$ (Table~\ref{table:groups}). The fixing approach that has the highest AAOD improvement is IN, which improves AAOD by 8.23\% on the CIFAR10S dataset and 33.66\% on the UTKFace dataset. This is because during the fixing process, IN ensures the features of each class are correctly represented in hidden layers, therefore mitigating bias. Layer-level approaches are not recommended for fairness-sensitive scenarios, as 3M degraded AAOD by 44.06\%, which is the worst performance among all approaches. This degradation stems from 3M's focus on local input optimizations, which primarily target samples from specific categories and thereby induce bias and fairness compromise. In particular, all neuron-level approaches preserve fairness on CIFAR10S. This is mainly because localizing specific neurons for repair in complex models reduces the likelihood of inducing unintended bias during the fixing process.

\finding{
Current fixing approaches generally minimally impact fairness. Neuron-level approach IN (fixing features) achieves the best fairness preservation on UTKFace and CIFAR10S. Layer-level approach 3M (verification-based) causes the most significant fairness degradation and is not recommended for fairness-sensitive scenarios.
}

Regarding NFR, layer-level approaches have an average NFR of 0.94\%, lower than that of neuron-level (1.72\%) and that of model-level (1.94\%) approaches. We observe that approaches with more extensive model modifications tend to induce more regression faults. For example, ET fixes 36.49\% of faults but introduces 12.39\% new regression faults, while GM induces no regression faults but hardly fixes any (except on MNIST). Among model-level approaches, finetuning-based ones (e.g., DR with an average NFR of 0.48\%) change training data distributions more slightly, thus inducing fewer regression faults than retraining-based approaches (e.g., AP with an average NFR of 2.25\%). Among neuron-level approaches, search-based ones (e.g., AR with an average NFR of 0.48\%, VERE with an average NFR of 0.06\%) induce fewer regression faults than training-based ones (e.g., IR with an average NFR of 4.05\%), likely because search-based approaches cause more conservative modifications. We also notice that the approaches with higher RR usually have worse NFR  (e.g., ET in group $\textit{a}$ for RR and group $\textit{g}$ for NFR in Table~\ref{table:groups}), emphasizing the need for balancing fixing effectiveness and backward compatibility.

\finding{
On average, layer-level approaches have lower NFR than neuron-level ones (0.94\% vs 1.72\%), which in turn outperform model-level ones (1.94\%). Approaches with more extensive modifications are more likely to adversely affect backward compatibility, with neuron-level training-based approaches performing worse than search-based ones and model-level retraining-based approaches underperforming finetuning-based ones.}

Overall, Table~\ref{table:groups} shows that no approach falls into group $\textit{a}$ across all metrics, indicating no single approach achieves superior fixing performance while improving accuracy and maintaining fairness, robustness, and backward compatibility. However, certain approaches demonstrate potential: AP achieves strong repair results on the small MNIST-LeNet5 model, improving robustness by 73.97\% and overall accuracy by 1.18\% with acceptable NFR (0.39\%). This is likely because AP leverages white-box information through multiple submodels for guided weight adjustment. The collective use of submodels aligns with the insights of ensemble learning~\cite{DBLP:journals/eaai/GanaieHMTS22}, highlighting opportunities for developing more balanced fix approaches preserving multiple properties.

\subsection{RQ4: Efficiency}

In practical applications, the time efficiency of each approach is a critical consideration. Thus, we evaluate the time cost of each approach. Notably, all reported experimental results are averaged over multiple trials. Table~\ref{table:time_cost} presents the results, where gray highlighting denotes the most efficient approach overall, and underlining indicates the most efficient approach within each category.

The results show that layer-level approaches are the most time-efficient, primarily due to their focus on specific layers (e.g., the fully-connected layers close to the outputs) for constraint-solving-based repairs. Among model-level approaches, HUDD and FM exhibit higher efficiency, attributed to their rapid data selection and augmentation processes compared to setting new training objectives (e.g., ET).
Neuron-level approaches like IN, IR, VERE, and AR demonstrate efficiency on different datasets. Notably, IN outperforms others on two datasets due to its use of training methods for fixing rather than search-based mechanisms. 
We also note that some approaches with high fixing effectiveness, e.g., AP, incur significant computational overhead (e.g., training 20 submodels to help adjust weights), requiring users to carefully consider their time cost before implementation.

\begin{table}[tbp]
\centering
\caption{Average Time Cost of Studied Approaches}
\label{table:time_cost}
\scalebox{0.7}{


\begin{tabular}{cc|ccccc}
\toprule
\textbf{Category}                      & \textbf{Approach} & MNIST  & UTKface & CIFAR10 & CIFAR10S & ImageNet \\
\midrule
\multirow{6}{*}{Model-level}  & AP       & 13h46m & 135h16m & 227h07m & 147h01m   & CORR.    \\
                              & ET       & 78h16m & CORR.   & CORR.   & 323h00m   & CORR.    \\
                              & FM       & 09m05s & 01h34m  & \ul{13m07s}  & \ul{02h43m}    & 25h32m   \\
                              & DR       & 01h10m & 05h25m  & 04h23m  & 10h41m    & 35h22m   \\
                              & HUDD     & \ul{01m05s} & \ul{41m18s}  & 06h46m  & 02h58m    & \ul{12h06m}   \\
                              & HR       & 17m12s      & 01h25m       & 01h22m       & 03h52m         & CORR.    \\
\midrule
\multirow{2}{*}{Layer-level}  & 3M       & 00m17s & \cellcolor{lightgray}\ul{00m23s}  & \cellcolor{lightgray}\ul{00m49s}  & \cellcolor{lightgray}\ul{00m47s}    & FAIL.    \\
                              & RA       & \cellcolor{lightgray}\ul{00m13s} & UNAD.   & 04m15s  & UNAD.     & UNAD.    \\

\midrule
\multirow{8}{*}{Neuron-level} & AR       & 01m12s & 01h18m  & 18m51s  & \ul{01m02s}    & 07h41m   \\
                              & CARE     & 04m28s & 11h04m  & 20m00s  & 12m29s    & 65h06m   \\
                              & NR       & 34m57s & 01h08m  & 06h27m  & FAIL.     & UNAD.    \\
                              & IR       & 17m00s      & \ul{09m43s}       & 36m05s      & 11m52s         & 144h30m        \\
                              & DTR      & UNAD.  & UNAD.   & 01h39m        & 01h24m         & 03h20m        \\
                              & GM       & 04h26m & 03h59m  & 38h50m  & 52h17m    & 02h25m   \\
                              & VERE     & UNAD.  & UNAD.   & \ul{03m51s}  & UNAD.     & UNAD.    \\
                              & IN       & \ul{00m35s} & 20m44s   & 12m55s    & 38m14s     & \cellcolor{lightgray}\ul{02h09m}       \\
                              \bottomrule
\end{tabular}
}

\begin{tablenotes}
\scriptsize
\item[*] Here are the abbreviations: CORR.: CORRUPTED; UNAD.: UNADAPTABLE; FAIL: FAILURE

\end{tablenotes}

\end{table}

\finding{
Layer-level approaches generally exhibit higher time efficiency compared to other approaches. In addition, HUDD and FM from the model-level, as well as IN, IR, VERE, and AR from the neuron-level, all demonstrate acceptable time efficiency. Although AP shows better fixing effectiveness, it also incurs significant computational overhead due to the utilization of white-box information.
}

\section{DISCUSSION}

\subsection{Key Takeaways for Industries}

\begin{description}[leftmargin=0pt]
\item[Recommendation of Fixing Approaches.]
Our findings confirm that no single method universally optimizes fixing performance while preserving all critical properties across datasets. Based on these results, we provide recommendations tailored to industrial developers' primary requirements:

\begin{itemize}[leftmargin=*]
\item \textbf{Overall Accuracy Preservation:} Based on Finding 2, we recommend using HUDD and FM (model-level) with small learning rates and VERE (neuron-level) to preserve overall accuracy.

\item \textbf{Extensive Faulty Behavior Correction:} Based on Finding 3, prioritizing using model-level approaches (e.g., AP, ET, HUDD) for correcting numerous misbehaving samples.

\item \textbf{Robustness Preservation:} Based on Finding 5, in tasks like autonomous driving or aerial systems where maintaining robustness is a high-priority non-functional requirement alongside correctness, using model-level AP and ET to maintain robustness against noise and adversarial examples is recommended.

\item \textbf{Fairness Preservation:}
Based on Finding 6, maintaining fairness is a high-priority non-functionality besides correctness for fairness-sensitive tasks (e.g., credit risk assessment, recruitment screening). Using IN (neuron-level) to preserve fairness is recommended. Avoid using DTR in fairness-sensitive scenarios like UTKFace to prevent bias induction.

\item \textbf{Backward Compatibility Preservation:} 
In safety-critical tasks relying on human operation (e.g., medical diagnosis, industrial control robots), based on Finding 7, using layer-level approaches (e.g., RA) is recommended. In addition, DR (finetuning-based, model-level) and AR\&VERE (search-based, neuron-level) are also recommended in their categories.

\item \textbf{Fixing Small Models:} Based on Finding 4, AP and ET offer superior fixing effectiveness for small-scale models if time permits.

\item \textbf{Fixing Large Models:} Based on Finding 4, finetuning approaches like FM and DR are recommended for large-scale models.

\item \textbf{Time-Constrained Repairs:} Based on Finding 8, layer-level approaches targeting output-adjacent layers are the most time-efficient, particularly for small models. However, if the models are too large to suit layer-level approaches, using neuron-level approaches such as AR and IN can be an alternative strategy.
\end{itemize}

\item[Applicability of Fixing Approaches.]

According to Finding 1, different approaches require different resources as inputs. 
For developers with \textbf{limited data resources}, self-contained approaches (i.e., AP, ET, GM, CARE, IN, AR) requiring only training data and model structures are recommended. If developers can acquire enough additional data, approaches like HUDD, IR, DR, FM, and HR are preferable. The following approaches typically demand more information:
1) \textbf{Specialized requirements}: DTR is recommended for scenarios where test data has concrete prioritization and risk levels due to its design tailored for prioritized/risk-weighted scenarios.
2) \textbf{Verification constraints}: If tasks require rigorous safety-critical verification, VERE, 3M, and RA are suitable.
Notably, NR is the only approach that requires historical versions for fixing. Developers should assess applicability of the approaches based on resource availability and task requirements.

\end{description}

\subsection{Implication for Researchers}
\begin{figure}[t!]

\includegraphics[scale=0.40]{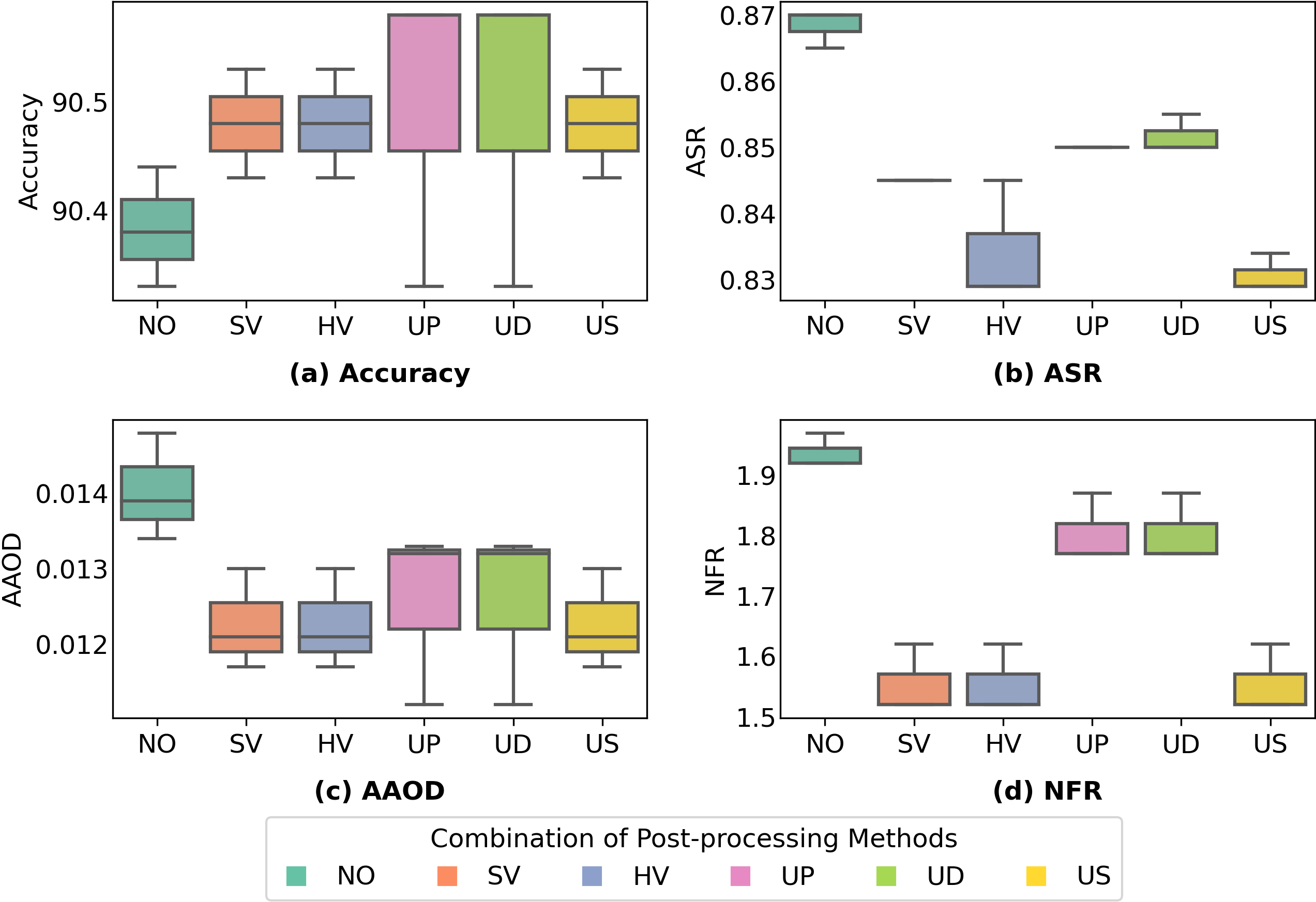}
\caption{Side Effects Mitigation with Post-processing}

\label{fig:diss}
\end{figure}

\begin{description}[leftmargin=0pt]
\item[Regularizing the Evaluation.]
Researchers should regularize the evaluation process. For example, if additional data is required for fixing, it should be selected from the repair set rather than the validation or test sets to avoid data leakage. The developers should also use the validation set instead of directly using the test set for evaluation to avoid overfitting.
Moreover, researchers should evaluate more properties during experiments, such as fairness, robustness, and backward compatibility. We commend the approach, IN~\cite{INNER}, as a good example of considering most of these metrics.

\item[Attention to Side-Effect Mitigation.] Findings 5-7 indicate that the fixing process can induce side effects on other properties. If developers aim to preserve these properties, they must mitigate such side effects. An effective strategy is to employ post-processing methods like model ensemble~\cite{voting} and uncertainty alignment~\cite{DBLP:conf/icse/LiZXYCCML23}.
We selected five state-of-the-art post-processing methods following existing work~\cite{you2025mitigating}: soft voting (SV), hard voting (HV), uncertainty alignment with perturbation (UP), dropout (UD), and temperature scaling (US). 
These methods fall into two categories: voting-based and uncertainty-based approaches. Voting-based approaches use majority vote for final decisions: SV averages predicted confidence scores from the original and fixed models, while HV selects the most frequent label (using the more confident model in case of ties).
Uncertainty-based approaches further consider the trustworthiness of each model. If test labels are unavailable, UP and UD estimate the uncertainty of prediction by injecting noise into inputs or using dropout on models, then align predictions accordingly. If test labels are available, US applies temperature scaling to calibrate model confidences and reduce their divergence.
We randomly sampled five subjects to cover all datasets and fixing categories: MNIST-ET, UTKFace-AP, CIFAR10-NR, CIFAR10S-3M, and ImageNet-HUDD for experimentation. Given the similar trends among them, we chose UTKFace-AP as an example for analysis.
Figure~\ref{fig:diss} shows that side effects on fairness, backward compatibility, and robustness (reflected by AAOD, NFR, and ASR) are mitigated, with lower values compared to no post-processing (denoted as NO in Figure~\ref{fig:diss}). Meanwhile, overall accuracy is maintained at a relatively high level. This result demonstrates the potential of post-processing methods. 
Although effective, these methods address symptoms rather than root causes.
Future research should clarify how side effects occur during fixing to enable proposing more mitigation strategies.

\item[Considering More Fixing Scenarios.]
In practice, industrial developers may need to fix incorrect behaviors while protecting the models' more important functionalities (e.g., those related to business interests or human safety) or properties (e.g., model security). This represents an important research direction. In addition, current approaches have limited effectiveness in fixing complex DL models (Finding 4). Given the increasing scale and complexity of DL systems, fixing these complex large-scale models poses a key challenge. There is an urgent need for scalable fixing approaches for large-scale architectures like LLMs and Agents.

\item[Enhancing Existing Fixing Strategies.]
Building on the limitations of current approaches, we identify several directions for improvement, including refining individual approaches and exploring their potential complementarity.
Training-based methods (e.g., ET, AP) often achieve high repair effectiveness but are prone to introducing regression faults. This issue may be alleviated by adding explicit objectives to control regressions (Finding 7).  Counterexample-based approaches (e.g., HUDD) tend to overfit due to the small number of counterexamples and could benefit from augmenting the set of positive examples (Finding 5).
In addition, some methods show complementarity at different levels. At the result level, HybridRepair and Apricot repair largely disjoint sets of samples, suggesting that they address different fault types. Combining their outputs using ensemble techniques such as majority voting with confidence-based weighting may enhance overall repair performance.
At the method level, data augmentation techniques modify training data distributions, while weight adjustment methods directly improve ill-trained model parameters. Integrating these approaches may provide more comprehensive repairs by addressing both data and neuron weaknesses. However, how to combine such strategies effectively while preserving their respective strengths remains an open question.

\end{description}

\subsection{Threats to Validity}

The threat to external validity lies in the selection of datasets. To ensure a fair and comprehensive evaluation of existing approaches, we followed prior work~\cite{ENN,VERE,Apricot,I-REPAIR,MVDNN,AIREPAIR} and adopted five widely used datasets and models for classification tasks. This standardization allows for consistent evaluation and facilitates extensive comparisons. In the future, we will explore the effectiveness of repair methods across different modalities (e.g., text, video), tasks (e.g., regression, generation), models (e.g., Transformer-based models or even LLMs with more complex structures).

The threat to internal validity mainly lies in the implementation of the studied approaches and the potential misjudgment of results. To mitigate the prior threat, we primarily used the open-source implementations from the corresponding papers when available. For other approaches, we double-checked their performance with the corresponding authors to ensure correct implementations and configurations. Moreover, we made all our experimental results open-source to facilitate replication and encourage future research.
To mitigate the latter threat, before drawing conclusions, we used the T-Test with Benjamini-Hochberg for p-value adjustment.

The threats to construct validity mainly lie in randomness. To reduce the threat of randomness, we repeated all approaches involving randomness three times~\cite{you2025mitigating} and calculated the average results in our study.


\section{CONCLUSION}
Numerous DL model fixing approaches have been proposed recently, yet many remain underutilized in the industry. To bridge this gap, we empirically investigated key industry concerns, including applicability, effectiveness, potential side effects, and time efficiency, by comprehensively evaluating 16 state-of-the-art approaches across five widely used datasets. 
Based on our findings, we provided recommended fixing approaches for different scenarios, highlighted implications for researchers to mitigate side effects and optimize existing fixing approaches, and thereby provided direction for future research and practical adoption.

\section*{Data Availability}
In the spirit of Open Science, we make the replication package with source code, detailed configuration settings, benchmark datasets, experiment scripts, and the experimental results publicly available on our homepage~\cite{homepage}.

\begin{acks}

This work was supported by the National Natural Science Foundation of China Grant Nos. 62472310 and 62322208, and by JSPS KAKENHI Grant No. JP23K28062.
\end{acks}

\bibliographystyle{ACM-Reference-Format}
\bibliography{sample-base}

\appendix









\end{document}